# Injecting Explainability and Lightweight Design into Weakly Supervised Video Anomaly Detection Systems

Wen-Dong Jiang, *Graduate Student Member, IEEE*, Chih-Yung Chang, *Member, IEEE*, Hsiang-Chuan Chang, Ji-Yuan Chen and Diptendu Sinha Roy, *Senior Member, IEEE*

***Abstract*—Weakly Supervised Monitoring Anomaly Detection (WSMAD) employs weak supervision learning to detect anomalies, crucial for monitoring in smart city. Existing multimodal methods often fail to meet real-time monitoring, and interpretability needs on edge devices due to their complexity. This paper introduces a Two-stage Cross-modal Video Anomaly Detection System named TCVADS. Using knowledge distillation and cross-modal contrastive learning, TCVADS enables efficient, accurate, and explainable anomaly recognition on edge devices. The system operates in two stages: coarse-grained rapid classification and fine-grained detailed analysis. In the first stage, TCVADS extracts features from video frames, inputting them into a time series analysis module that serves as the teacher model. Knowledge distillation then transfers insights to a simplified convolutional network (student model) for binary classification. The second stage, triggered upon anomaly detection, employs a fine-grained multi-class classification model. It uses CLIP for cross-modal contrastive learning with text and images, achieving more refined and interpretable classification through specially designed triplet textual relationships. Experimental results demonstrate TCVADS significantly outperforms existing technologies in model performance, detection efficiency, and interpretability, contributing to monitoring applications in smart cities.**

***Index Terms*—Video Anomaly Detection, Intelligent monitoring, Explanation, Smart City.**

This work was supported by the National Science and Technology Council of Taiwan, Republic of China, under grant number NSTC 112-2622-E-032-005. (Corresponding author: Chih-Yung Chang.)

Wen-Dong Jiang is currently pursuing a Ph.D. degree with the Department of Computer Science and Information Engineering, Tamkang University, New Taipei 25137, Taiwan. (e-mail: 812414018@o365.tku.edu.tw).

Chih-Yung Chang is with the Department of Computer Science and Information Engineering, Tamkang University, New Taipei 25137, Taiwan. (email: cychang@mail.tku.edu.tw).

Hsiang-Chuan Chang is with the Department of Transportation Management, Tamkang University, New Taipei 25137, Taiwan. (149190@o365.tku.edu.tw)

Ji-Yuan Chen is currently pursuing a Ph.D. degree with The Faculty of Engineering and Information Technology, The University of Melbourne, Parkville VIC 3052, Australia. (email: jiyuanc1@student.unimelb.edu.au).

Diptendu Sinha Roy is with the Department of Computer Science and Engineering, National Institute of Technology, Shillong, 793003, India (e-mail: diptendu.sr@nitm.ac.in).

## I. INTRODUCTION

Weakly supervised Monitoring anomaly detection (WSMAD) has emerged as a research hotspot in the realm of IoT-based computer vision. Its growing importance is driven by its wide-ranging potential applications in intelligent surveillance [1], smart city [2], and traffic monitoring [3]. WSMAD employs weakly supervised learning methods to detect anomalous behaviors in videos, enabling models to be trained without the need for explicitly labeled anomalies.

Previous WSMAD research typically relied on a single visual modality [6] [38] to detect abnormal behaviors. However, visual cues alone may fail to accurately capture potential anomalies. For instance, in a school environment, verbal abuse by a teacher towards a student is difficult to detect through images alone. To improve the accuracy of anomaly detection systems, current WSMAD research frameworks often combine multi-modal information. This includes gathering data from images, audio, and other sources. For example, in smart monitoring systems [32] [27], data from sound sensors, environmental sensors, and even social media information is used to enhance the detection of abnormal behaviors. However, these methods rely heavily on external devices, which can create significant challenges during real-world deployment. These challenges include high equipment procurement costs, difficulties in sensor deployment, and substantial data transmission delays. Additionally, as the number of modalities increases, the issue of interpretability also becomes a complex problem [31] [33].

In recent years, some researchers [34] [4] have tried to use cross-modal methods to solve challenges in single-image anomaly detection. This method typically used information from one modality to infer or supplement information from another modality, reducing reliance on external devices. However, using cross-modality for anomaly detection in video content still faced two major challenges. The first challenge was the high computational complexity associated with cross-modal methods. These methods required processing data from different modalities and establishing cross-modal connections. This extensive computational workload reduced the real-time performance of the system, making it challenging to make quick judgments in practical applications. The second challenge was the lack of in-depth cross-modal analysis. Existing cross-modal detection methods often remained at a shallow level of data fusion, making it difficult to fully explore the connections between different modalities. This limitation led to a model with a limited ability to detect abnormal behavior effectively.

In terms of immediacy, the current WSMAD system framework [24] [25] [26] typically used pre-trained visual models (such as C3D [5], I3D [6], ViT [7]) to extract video frame features. These features were then fed into temporal models like Transformers and GCNs for local and global analysis. Finally, the model training was conducted based on binary or multi-class classifiers using multi-instance learning (MIL) to detect abnormal event confidence. However, the high computational complexity associated with embedding features poses challenges for practical applications. Table I illustrates the computational complexity of this architecture: Herein, $n$ represents the length of the input sequence, $d$ represents the feature dimension, $m$ represents the number of edges of GCN, $w$ represents the window size, where $w < n$ ; $h$ is the number of heads of Transformer multi-head attention, and $k$ represents the number of layers of GCN.

TABLE I

COMPLEXITY STATISTICS OF THE WSMAD SYSTEM

| Model | The Complexity |
|---|---|
| Transformer multi-Attention | $C_A(T)_{window} = O(h \cdot n \cdot w \cdot d)$ |
| Transformer Fully Connect | $CF(T) = O(n \cdot d^2)$ |
| GCN | $C(GCN) = O(k \cdot m \cdot d)$ |
| $(Total) = C_A(T)_{window} + CF(T) + C(GCN)$ $= O(h \cdot n \cdot w \cdot d) + O(n \cdot d^2) + O(k \cdot m \cdot d)$ | |

Therefore, reducing computational complexity and improving immediacy is crucial for efficient feature extraction and timing analysis in WSMAD. Additionally, leveraging visual-verbal connections can significantly enhance anomaly detection performance, especially considering text's superiority over images in unstructured data scenarios. This approach could lead to more comprehensive and effective anomaly detection systems.

Over the past three years, visual-linguistic pre-trained (VLP) models, such as CLIP, have shown remarkable performance in computer vision tasks. CLIP [8], for instance, aligns images and text using contrastive learning, ensuring that matching descriptions are closer while mismatching ones are separated. It leverages hundreds of millions of image-text pairs from the web, showcasing its robust representation learning capabilities and strong visual-language correlation. The success of CLIP has led to the emergence of more task-specific models that excel in various visual tasks, achieving unprecedented performance levels. However, it's worth noting that CLIP and similar models have primarily focused on the image domain [9][10], and their applicability in complex WSMAD tasks requires further investigation and study. Moreover, how to ensure that these models are interpretable is also a difficult problem that needs to be overcome. Integrating such advanced Vision-Language Pre-training (VLP) models into WSMAD frameworks could potentially enhance anomaly detection performance by leveraging the rich visual-verbal connections they offered.

To address the challenges of computational complexity and real-time performance in weakly supervised video anomaly detection systems, the existing solutions mainly include the following approaches: single-modality optimization methods [1][2], which aimed to improve detection efficiency by optimizing features from a single modality (such as visual data); and more efficient model architectures [3], which reduced computational costs by designing lightweight network structures. However, these methods have significant limitations: they cannot simultaneously improve detection accuracy and significantly reduce computational resources. Therefore, we adopt multimodal methods as our solution. By integrating information from multiple modalities (such as visual and text), multimodal systems not only enhance detection accuracy but also reduce feature redundancy and increase system robustness, offering unique advantages.

This paper proposes a Two-stage Cross-modal Video Anomaly Detection System, named TCVADS. In the first stage, TCVADS employs an enhanced MobileNet module for rapid feature extraction. Then, it analyzes the time series both locally and globally using the Enhanced Recurrent Weighted Average Kernel Unitary Vector (RWKV) module, replacing the conventional Transformer and GCN combination. This approach enables Enhanced RWKV module to learn time-sequence features while reducing the computational complexity to $O(n \cdot d)$ . Additionally, TCVADS transfers the key knowledge of RWKV to a three-layer CNN, referred to as the Quick Anomaly Classification Module (QACM), through knowledge distillation for fast-response coarse-grained detection. Upon detecting potential anomalies, TCVADS proceeds to the second stage. In this stage, it applies an improved CLIP module along with learnable prompts for precise fine-grained detection in image-text alignment. Experimental results indicate that TCVADS maintains high detection accuracy while significantly reducing computational complexity, thereby improving the system's responsiveness and interpretability. The contributions of this study can be summarized in three key aspects:

1. **Novel Framework TCVADS:** The novel framework TCVADS is presented for video anomaly detection using a two-stage approach involving visual classification and visual-linguistic alignment.
2. **Rapid and Video Anomaly Detection:** The proposed RWKV-based module, compared to architectures using Transformer and GCN, only requires a computational complexity of $O(n \cdot d)$ to effectively capture temporal dependencies. Furthermore, through knowledge distillation, the learned knowledge is transferred to a simple CNN network, showing faster speeds during both training and trial phases than in previous work.
3. **Innovative Learnable Prompt Engineering:** This paper introduces an innovative learnable prompt engineering method aimed at enhancing the precision and interpretability of image-text matching in contrastive learning. By integrating Clip's original text with fine-grained detection of abnormal videos and using learnable prompts as inputs for the text encoder, this approach not only improves the efficacy of contrastive learning but also enhances its interpretability.

The remainder of the paper is organized as follows. Section II discusses and compares previous relevant studies. Section III describes the Assumptions and problem formulation in detail. Section IV details the TCVADS proposed in this paper. Section V provides the experiments and performance evaluation. The conclusion is discussed in Section VI.

## II. RELATED WORK

This chapter presents a review of some relevant studies in Video Anomaly Detection. These studies are categorized into two groups: Weakly supervised video anomaly detection and using visual language pre-training model to detect anomalies.

TABLE II

COMPARISON WITH OTHER WORK

| Related Works | Model | Fine-Grained and Coarse-grained | Two-Stage | Cross-modal | Real-Time | Accurate Prompt |
|---|---|---|---|---|---|---|
| [6] (2016) | C3D | Coarse-grained | × | × | × | × |
| [5] (2018) | I3D | Fine-Grained | × | × | × | × |
| [14] (2020) | HD-Net | Coarse-grained | × | × | × | × |
| [12] (2021) | RTFM | Coarse-grained | × | ✓ | × | × |
| [38] (2022) | OCSVM | Coarse-grained | × | × | ✓ | × |
| [15] (2022) | AVVD | Fine-Grained and Coarse-grained | × | × | × | × |
| [8] (2022) | CLIP | Coarse-grained | × | ✓ | × | × |
| [13] (2023) | DMU | Coarse-grained | × | × | × | × |
| [3] (2023) | UMIL | Coarse-grained | × | ✓ | × | × |
| [4] (2023) | CLIP-TSA | Coarse-grained | × | ✓ | × | × |
| [21] (2024) | VadCLIP | Fine-Grained and Coarse-grained | × | ✓ | × | ✗ |
| [36] (2024) | AnomalyCLIP | Fine-Grained | × | ✓ | × | ✓ |
| [37] (2024) | STPrompt | Fine-Grained and Coarse-grained | × | ✓ | × | ✓ |
| **Ours** | **TCVADS** | Fine-Grained and Coarse-grained | ✓ | ✓ | ✓ | ✓ |

### *A. Weakly supervised video anomaly detection*

In WSMAD, *Sultani et al.* [5] and *Hasan et al.* [6] first proposed a deep multi-instance learning model for anomaly detection. The study treated the frame of the video as a package, in which multiple paragraphs were regarded as instances to enhance the detection performance of the model. Subsequent research was based on self-attention or the Transformer or GCN model to model the temporal and contextual relationships in video content. *Zhong et al*. [11] proposed a GCN-based method for modeling feature similarity and temporal consistency between video segments. *Tian et al.* [12] used a self-attention network to capture the global temporal context of videos. *Wu et al.* [13] proposed a global and local attention module to capture temporal dependencies in videos to obtain more expressive embeddings. *Ji* and *Lee* [38] proposed OCSVM for anomaly detection. The above methods, only detect video frames were abnormal, which only provided a solution for a binary problem. While *Wu et al.* [14] proposed a fine-grained WSMAD method to distinguish different types of abnormal frames. Recently, the CLIP model has also attracted great attention in the VAD field. Based on the visual features of CLIP, *Lv et al.* [3] proposed a multi-instance learning framework called UMIL to learn unbiased anomaly features that improved WSMAD task performance. *Joo et al.* [4] used the visual features of CLIP to efficiently extract discriminative representations, long-term, and short-term temporal dependencies through temporal self-attention, and nominated interesting clips.

All the above methods were based on the classification paradigm and detected abnormal events by predicting the probability of abnormal frames. However, this classification paradigm did not fully exploit the semantic information of text labels. In addition, multi-modal fusion is still difficult in practical applications and many methods cannot fully exploit the semantic connections between text and visual features. Finally, although Transformer or GCN-based architectures are effective in modeling complex temporal relationships, they make the real-time performance of video anomaly detection challenging.

### *B. Visual language pre-training*

Significant progress had been made in visual-language pre-training, which aimed to learn the semantic correspondence between vision and language by pre-training on large-scale data. As one of the most representative works, CLIP had demonstrated excellent performance in a series of visual-linguistic downstream tasks. For example, *Zhou et al.* [15] proposed an improved image classification method based on CLIP; *Mokady et al.* [16] made a breakthrough in image subtitle generation; *Zhou et al.* [17] applied the contrastive learning method to target detection; Yu et al. [18] made achievements in the field of scene text detection. *Rao* and *Zhou* et al. [19] conducted research on dense prediction.

Recently, some follow-up work attempted to utilize such pre-trained models for applications in the video field. For example, *Luo et al.* [20] proposed CLIP4Clip, a method aimed at transferring the knowledge of CLIP models to the field of video-text retrieval; *Wu et al.* [21] proposed a simple but powerful baseline that efficiently adapted pre-trained image-based visual-language models to leverage their powerful capabilities in general video understanding and applied them to video-level downstream weakly supervised video anomaly detection. *Zhou et al.* [36] proposed an effective zero-shot anomaly defect detection method using a pre-trained CLIP model. This approach allowed anomaly detection without the need for specific training data by leveraging CLIP's ability to match visual features with textual descriptions. *Wu et al.* [37] introduced a CLIP-based three-branch architecture to address classification and localization in weakly supervised anomaly detection. This architecture was designed to simultaneously tackle the problems of coarse-grained detection, fine-grained detection, and localization in weakly supervised violence detection. However, most of these CLIP-based methods are designed using only text or image encoders, and the prompt words in the text encoder are mostly designed manually or through simple prompt engineering, making it difficult to fully utilize the advantages of contrastive learning.

Table II summarizes the aforementioned methods and compares them with the proposed model in terms of Fine-Grained and Coarse-grained, Two-Stage, Cross-modal, Real-Time, and Accurate Prompt.

## III. ASSUMPTIONS AND PROBLEM FORMULATION

This section introduces the assumptions and problem statements of this study. Given a video $V$ with a duration of $t$, this paper aims to identify whether or not the anomaly in $V$. Let $\hat{V}_i \in V$ and $V_i \in V$ denote non-anomaly and anomaly videos, respectively. Let video $V = \{\Phi_1, \Phi_2, \dots, \Phi_n\}$ be divided into $n$ equal-length segments, each containing non-anomaly, anomaly or a combination of both. Each segment $\Phi_i = \hat{V}_i \cup V_i$ might comprise both $\hat{V}_i$ and $V_i$ behaviors.

Let $C = \{c_1, c_2, \dots, c_m, \hat{c}\}$ denote all possible classes, where $c_i$ denotes the $m$ anomaly classes for $1 \le i \le m$ and $\hat{c}$ denotes the non-anomaly class. Let $L = \{l_1, l_2, \dots, l_m, \hat{l}\}$ be the set of labels, where $l_i$ is the corresponding label of class $c_i$ for $1 \le i \le m$ and $\hat{l}$ denote the label of non-anomaly class $\hat{c}$. This study assumes that each $V_i$ falls into a specific anomaly category $c_j$. Consider an anomaly detection mechanism $\mathcal{M}$ which aims to detect the occurrence of an anomaly event. Let $R^{\mathcal{M}}$ be a segment of video which is detected as the anomaly event by applying mechanism $\mathcal{M}$. That is, $R^M$ can be represented as $R^M = \{\hat{R}_1^M, R_1^M, \hat{R}_2^M, R_2^M \dots\dots, \hat{R}_n^M, R_n^M\}$.

Let $\delta_i$ be a Boolean variable that indicates whether or not a given video segment $\tilde{R}_i$ actually contains an anomaly event. That is $\delta_i = \begin{cases} 1, & \tilde{R}_i \in c_j \\ 0, & \tilde{R}_i \in \hat{c} \end{cases}$. Let $\theta$ be the prediction threshold. Let $\delta_i^M(\theta)$ denote whether or not the result of video segment $\tilde{R}_i$ predicted by $M$ contains anomaly event. Let probability $P_i$ denote the prediction score of $\tilde{R}_i \in R^M$. That is $\delta_i^M(\theta) = \begin{cases} 1, & if\ P_i \ge \theta \\ 0, & otherwise \end{cases}$.

Let $TP_i$, $TN_i$, $FP_i$ and $FN_i$ represent True Positive, True Negative, False Negative and False Positive respectively, of the prediction result of input $\tilde{R}_i$ by applying mechanism $M$. The values of $TP_i$, $FP_i$, $TN_i$ and $FN_i$ can be calculated using $TP_i = \delta_i \times \delta_i^M(\theta)$, $TN_i = (1-\delta_i) \times (1-\delta_i^M(\theta))$ $FP_i = (1-\delta_i) \times (1-\delta_i^M(\theta))$ and $FN_i = \delta_i \times (1-\delta_i^M(\theta))$, respectively.

Let $TP$, $TN$, $FP$ and $FN$ represent the cumulative True Positive, True Negative, False Positive, and False Negative, respectively, of the prediction results by applying mechanism $M$ to all segments $\tilde{R}_i \in R$, for $1 \le i \le n$. The value of $TP$, $TN$, $FP$, and $FN$ can be calculated by $TP = \sum_{i=1}^n TP_i$, $TN = \sum_{i=1}^n TN_i$, $FP = \sum_{i=1}^n FP_i$ and $FN = \sum_{i=1}^n FN_i$ respectively.

Let $\mathscr{p}^M$, $\mathcal{R}^M$ denote the Precision and Recall of the predictions by applying mechanism $M$ to predict a given video $V$. The values of $\mathscr{p}^M$, $\mathcal{R}^M$ can be calculated by $\mathscr{p}^M = \frac{TP}{TP+F}$ and $\mathcal{R}^M = \frac{TP}{TP+FN}$ respectively.

### A. *Average Precision (AP)*

Let $\theta_j$ denote the $j$-th prediction threshold, let $\mathcal{AP}^{\mathcal{M}}$ denote the *Average Precision* of mechanism $\mathcal{M}$. The $\mathcal{AP}^{\mathcal{M}}$ can be calculated by Exp. (1):

$$\mathcal{AP}^{\mathcal{M}} = \sum_{j=1}^{n-1} \left(\mathcal{R}(\theta_{j+1})^{\mathcal{M}} - \mathcal{R}(\theta_j)^{\mathcal{M}}\right) \times \mathscr{p}(\theta_{j+1})^{\mathcal{M}}$$

$$= \sum_{j=1}^{n-1} \left( \frac{TP(\theta_{j+1})}{TP(\theta_{j+1}) + FN(\theta_{j+1})} - \frac{TP(\theta_j)}{TP(\theta_j) + FN(\theta_j)} \right) \times \frac{TP(\theta_{j+1})}{TP(\theta_{j+1}) + FP(\theta_{j+1})}$$

$$= \sum_{j=1}^{n-1} \left( \frac{\sum_{i=1}^n \delta_i \times \delta_i^M(\theta_{j+1})}{\sum_{i=1}^n \delta_i \times \left(1 - \delta_i^M(\theta_{j+1})\right)} - \frac{\sum_{i=1}^n \delta_i \times \delta_i^M(\theta_j)}{\sum_{i=1}^n \delta_i \times \left(1 - \delta_i^M(\theta_j)\right)} \right) \times \frac{\sum_{i=1}^n \delta_i \times \delta_i^M(\theta_{j+1})}{\sum_{i=1}^n (1-\delta_i) \times \delta_i^M(\theta_{j+1})}. \quad (1)$$

Similar to previous works [14], [21], [38], the first objective of this paper is to develop mechanism $M^{best}$ that satisfies Exp. (2):

**First Objective in XD-Violence:**

$$M^{best} = arg \underset{M \in \mathcal{M}}{Max}(\mathcal{AP}^{\mathcal{M}}) \quad (2)$$

### *B. Area Under the Curve (AUC) and AUC-ano*

Let $\mathcal{A}_M$ denote the *AUC* of mechanism *M.* The value of $\mathcal{A}_M$ can be calculated by Exp. (3)

$$\mathcal{A}_M = \int_0^1 TPR(\theta)\, d\big(FPR(\theta)\big)$$

$$= \int_0^1 \frac{TP(\theta)}{TP(\theta) + FN(\theta)} d\left(\frac{FP(\theta)}{FP(\theta) + TN(\theta)}\right)$$

$$= \int_0^1 \frac{\sum_{i=1}^n \delta_i \times \delta_i^M(\theta)}{\sum_{i=1}^n \delta_i \times \left(1 - \delta_i^M(\theta)\right)} d\left(\frac{\sum_{i=1}^n (1-\delta_i) \times \delta_i^M(\theta)}{\sum_{i=1}^n (1-\delta_i) \times \left(1 - \delta_i^M(\theta)\right)}\right). \quad (3)$$

Similar to previous works [14], [21], [36], [38], the second objective of this paper is to develop mechanism $M^{best}$ that satisfies Exp. (4):

**Second Objective in UCF-Crime:**

$$M^{best} = arg \underset{M \in \mathcal{M}}{Max}(\mathcal{A}_M) \quad (4)$$

Let $\delta_j$ denote the indicator variable for the $j$-th instance, where $\delta_j = 1$ indicates an anomaly, and $\delta_j = 0$ indicates a normal instance. Let $\prod(\delta_i^M > \delta_j^M)$ denote an indicator function that equals 1 if the anomaly score of instances $i$ is greater than that of instance $j$, and 0 otherwise. Let $\mathcal{A}no_M$ denote the Ano-AUC of mechanism *M.* The value of $\mathcal{A}no_M$ can be calculated by Exp. (5):

$$\mathcal{A}no_M = \frac{\sum_{i=1}^n \sum_{j=1}^n \delta_i \times (1-\delta_j) \times \prod(\delta_i^M > \delta_j^M)}{\sum_{i=1}^n \delta_i \times \sum_{j=1}^n (1-\delta_j)}, \quad (5)$$

Similar to previous works [14], [21], [36], [38]. The third objective of this paper is to develop mechanism $M^{best}$ that satisfies Exp. (6):

**Third Objective in UCF-Crime:**

$$M^{best} = arg \underset{M \in \mathcal{M}}{Max}(\mathcal{A}no_M) \quad (6)$$

This section introduced the assumptions and problem formulation of this paper. The next section will introduce the proposed TCVADS.

## IV. TCVADS SYSTEM

This section introduces the proposed TCVADS system in detail. The primary goal of the proposed TCVADS is to develop an interpretable anomaly detection system named TCVADS, which can operate on edge devices, identify the temporal and spatial features of anomalous events, and accurately and quickly classify and explain them.

The proposed TCVADS mainly consists of two stages. The primary goal of the first stage is to construct and train a binary classification model that extracts features from abnormal videos and allows for quick binary classification, rapidly determining whether the input video is normal or abnormal. If the input video segment is potentially abnormal, it then proceeds to the second stage of fine-grained judgment to identify the specific type of abnormal situation. The fine-grained precise comparison in the second stage can perform a multi-classification task on the input video, determining whether the input video belongs to one of the six categories, such as fighting, threatening, shooting, etc. Fig. 1 shows the specific process of TCVADS. The upper and lower blocks denote the Coarse-Grained and Fine-Grained stages, respectively. The following is the entire implementation process:

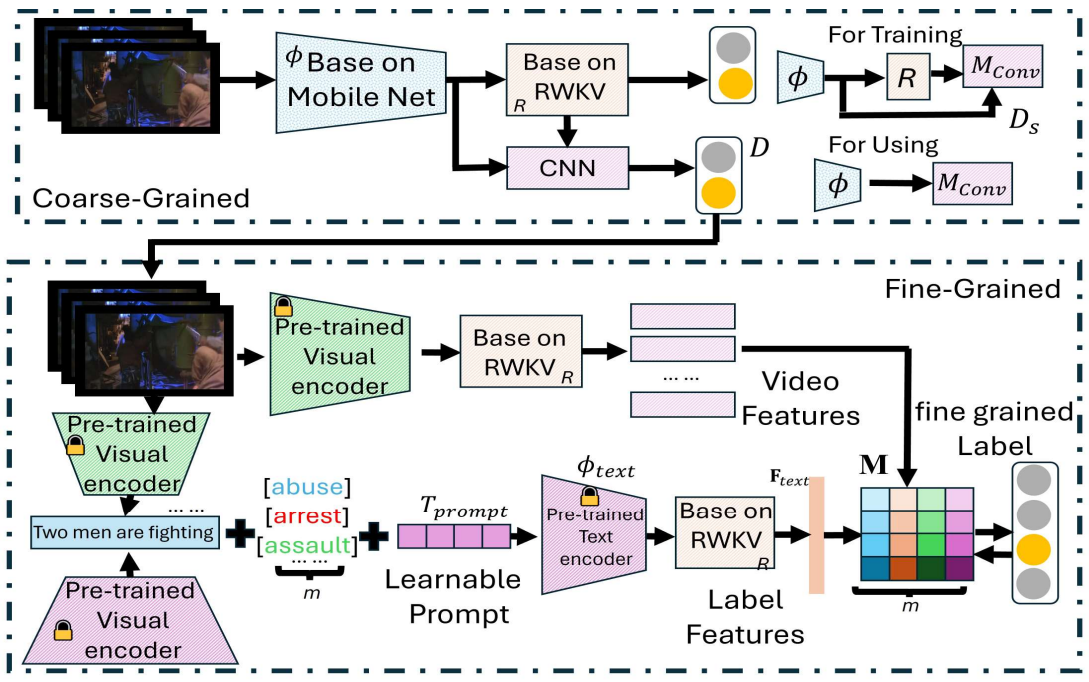

Fig. 1 Design flow chart of TCVADS.

### A. *TCVADS Coarse-grained detection*

The first stage focuses on developing and training a binary classification model that can effectively extract features from temporally anomalous videos. This is achieved through a knowledge distillation training phase, which enables the model to learn and capture the essential characteristics of anomalous behavior in video sequences. To achieve this, the proposed approach first trains a complete model, the Anomalous Feature Extraction and Detection (*AFED*) module, capable of extracting temporal features from anomalous videos. The fast binary anomaly classification model, Quick Anomaly Classification Module (*QACM*), which is the actual model intended to be used, extracts its important features through knowledge distillation from the *AFED* module to complete the training of the *QACM* module.

Fig. 2 illustrates this design, where the complete *QACM* module is primarily modified from MobileNet, followed by a model with temporal anomaly binary classification capabilities, which is mainly adapted from the RWKV model. Initially, labeled videos are used to train the parameters of the *AFED* module, enabling it to classify anomalous images. Subsequently, the next step is to perform knowledge distillation on the *AFED* model, allowing the QACM module to quickly learn to extract temporal features and rapidly determine whether a video is anomalous.

(1) *AFED* module

As shown in Fig. 2, the *AFED* module includes two stages. The first stage aims to extract interpretable features from each frame of the input video segment. The second stage focuses on capturing the time-series patterns from the frame features in a more lightweight manner. The advantage of the two-stage design of the AFED module lies in its ability to not only improve detection accuracy by effectively handling both spatial and temporal anomalies but also ensure the scalability of the module. This allows it to process longer video sequences or larger datasets without a significant increase in computational resource consumption.

In the first step, let $\boldsymbol{F} = \{f_1, f_2, \dots, f_x\}$ denote the $x$ input video frames, where $f_i$ represents the $i$-th frame. In the first stage, the feature extraction Module aims to extract the features from each $f_i \in \boldsymbol{F}$. The proposed feature extraction Module is a redesigned MobileNet [30], with deconvolution added to each convolution module to enhance the interpretability of feature extraction.

In many deep learning applications, the black-box nature of models makes it difficult to explain their decision-making processes, which is particularly worrisome in sensitive scenarios such as video violence detection. This paper introduces deconvolution operations aimed at visualizing and explaining the generation process of each feature map, thereby providing insights into the inner workings of the model and making it more transparent and trustworthy for users and regulators.

Let $\boldsymbol{C}$ and $\mathbf{D}$ denote the convolution and deconvolution operations, respectively. The feature extraction process for each frame $f_i \in F$ can be described by $f_i = \sigma(\mathbf{W}_D \cdot \sigma(\mathbf{W}_C \cdot f_t + \mathbf{b}_C) + \mathbf{b}_D)$. Where $\mathbf{W}_C$ and $\mathbf{W}_D$ are the weight matrices for convolution and deconvolution operations, respectively, $\mathbf{b}_C$ and $\mathbf{b}_D$ are bias vectors, and $\sigma$ is the nonlinear activation function.

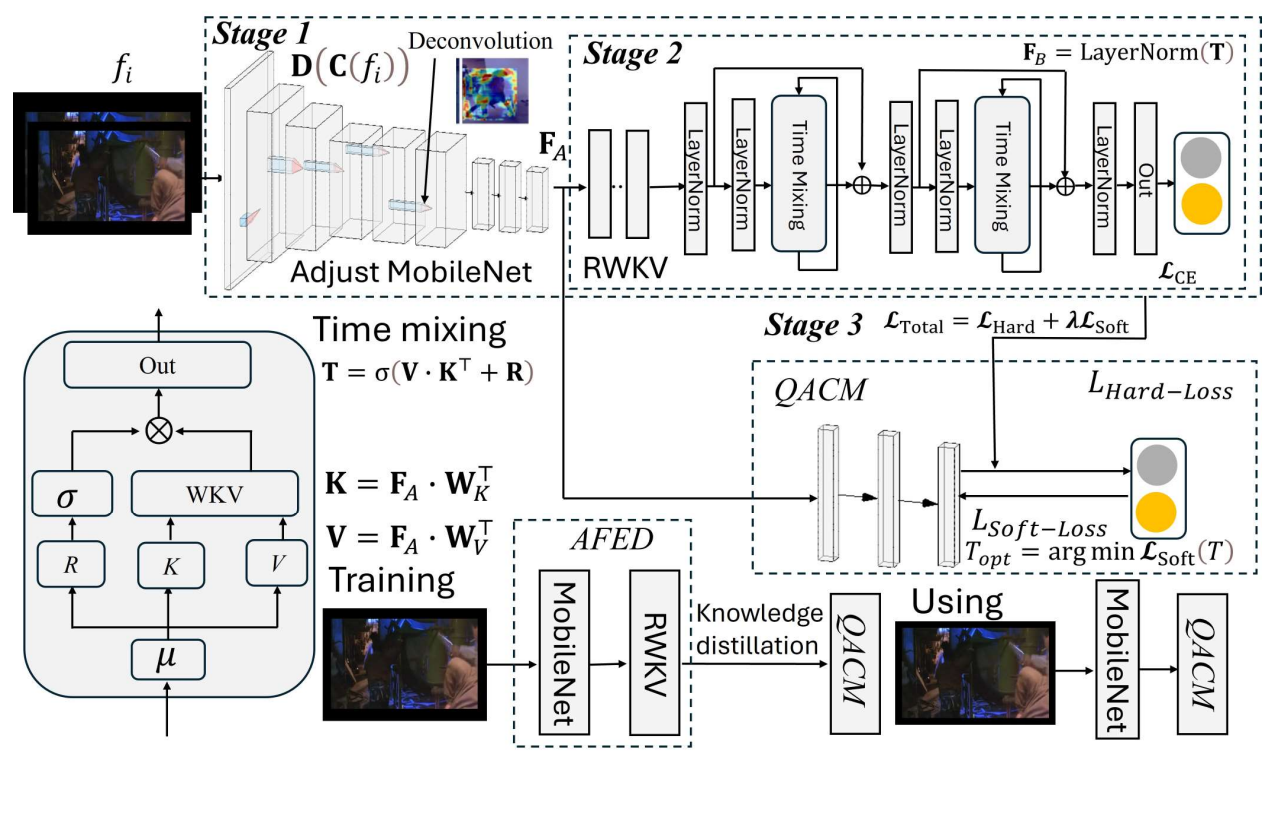

Fig.2 The TCVADS Coarse-Grained Working Flow.

***Proof: Enhanced Interpretability Property:***

The proposed Adjust MobileNet enhances the Interpretability of feature extraction. To more clearly demonstrate this point, the impact of input pixels on the feature map can be quantified by analyzing the gradients. Specifically, by calculating the partial derivatives of the convolution and deconvolution operations with respect to the input, it is possible to reveal how each input pixel affects the output feature map. This approach not only helps in quantifying this impact but also intuitively illustrates the internal workings of the model.

The gradient of the convolution feature map $\boldsymbol{F}_C$ with respect to the input $f_t$ is given by $\frac{\partial \boldsymbol{F}_C}{\partial f_t} = W_C^\top \cdot diag\big(\sigma'(\boldsymbol{W}_C \cdot f_t + \boldsymbol{b}_C)\big)$, where $\sigma'$ denotes the derivative of $\sigma$. For the deconvolution feature map $\mathbf{F}_A$, the gradient is expressed $\frac{\partial \boldsymbol{F}_A}{\partial f_t} = W_D^\top \cdot diag(\sigma'(\boldsymbol{W}_D \cdot I_t + \boldsymbol{b}_C) + \boldsymbol{b}_D) \cdot W_C^\top \cdot diag\big(\sigma'(\boldsymbol{W}_C \cdot f_t + \boldsymbol{b}_C)\big)$.

Assume that $\sigma'$ is predominantly non-zero and the input $I_t$ contains violence. The value of $\sigma'$ can be simplified as $\sigma' \approx 1$. As a result, we have $\frac{\partial \boldsymbol{F}_C}{\partial f_t} \approx W_C^\top, \frac{\partial \boldsymbol{F}_A}{\partial f_t} \approx W_D^\top \cdot W_C^\top$.

Suppose $\boldsymbol{W}_D$ is designed such that the squared Frobenius norm of $\boldsymbol{W}_D$ is greater than unity ($\|\boldsymbol{W}_D\|_F^2 > 1$). Exp. (7) gives a comparison between the squared Frobenius norms of two gradients.

$$\left\|\frac{\partial \mathbf{F}_A}{\partial f_t}\right\|_F^2 > \left\|\frac{\partial \mathbf{F}_C}{\partial f_t}\right\|_F^2. \tag{7}$$

This inequality substantiates that the gradient magnitude of $\mathbf{F}_A$ is greater than that of $\mathbf{F}_C$, implying that the deconvolution layer significantly enhances the influence of input pixels on the output feature map, thus increasing interpretability.

After completing the interpretable feature extraction, the next step is to capture time-series patterns from the frame features in a more lightweight manner. The design of the time-series analysis in the AFED module is modified based on the RWKV model [29], enabling more lightweight local and global analysis of the time-series data. In the traditional RWKV model [29], RWKV mainly consists of two submodules: Time Mixing and Channel Mixing. The Time Mixing module captures dependencies in the time dimension, while the Channel Mixing module integrates information between different channels. By borrowing the weight-sharing mechanism of RNNs, the attention module of Transformers has been improved to achieve infinite context modeling and low-latency inference in text tasks.

However, in video anomaly detection tasks, the local temporal modeling of Channel Mixing may be unnecessary. Firstly, the modeling of spatial features has already been completed in the feature extraction module. To further enhance RWKV's performance in long-sequence modeling tasks, this paper proposes an enhanced RWKV model, which includes two Time Mixer modules to replace the original RWKV architecture, aiming for more efficient time series modeling. By cleverly combining the advantages of RNN and Self-Attention, the proposed enhanced RWKV model can capture global temporal dependencies with lower computational costs. This ensures that the model's complexity remains at $O(n \cdot d)$ during operation. The following is the specific operational process of the enhanced RWKV model:

The enhanced RWKV module first calculates the Key and Value of the input feature vector $\boldsymbol{F}_A$. Assume the input feature matrix $\mathbf{F}_A \in \mathbb{R}^{n \times d}$, where $n$ is the number of time steps and $d$ is the feature dimension. The notations $\mathbf{W}_K$ and $\mathbf{W}_V$ denote the weight matrices for the Key and Value, respectively. They are both elements of $\mathbb{R}^{d \times d}$. The matrix multiplication task is divided into $p$ subtasks, with each processor core responsible for a portion of the computation. The matrix $\mathbf{F}_A$ is partitioned into $p$ submatrices $\mathbf{F}_{A,i}$, and multiply the weight matrices $\mathbf{W}_K^\top$ in parallel, obtaining the submatrix $\mathbf{K}_i$, as shown in $\boldsymbol{K}_i = \boldsymbol{F}_{A,i} \cdot \boldsymbol{W}_K^\top, \boldsymbol{V}_i = \boldsymbol{F}_A \cdot \boldsymbol{W}_V^\top$.

Combine all the $\boldsymbol{K}_i$ matrices to form the final result matrices $\boldsymbol{K}$ and $\boldsymbol{V}$. With $\boldsymbol{p}$ processor cores, the optimized total computational complexity is $O(n \cdot d)$. Next, in the time mixing module, the key operation is time mixing which combines the Key and Value for global and local analysis. The operation of time mixing operation $\mathbf{T}$ is shown in Exp. (8):

$$\begin{aligned}\mathbf{T} &= \sigma(\mathbf{V} \cdot \mathbf{K}^\top + \mathbf{R}) \\ &= \sigma\left(\begin{bmatrix}\mathbf{V}_1^\top \mathbf{K}_1 & \cdots & \mathbf{V}_1^\top \mathbf{K}_d \\ \vdots & \ddots & \vdots \\ \mathbf{V}_n^\top \mathbf{K}_1 & \cdots & \mathbf{V}_n^\top \mathbf{K}_d\end{bmatrix} + \mathbf{R}\right),\end{aligned}$$

where $\mathbf{R}$ is the memory unit matrix. Similarly, the computational complexity for Exp. (16) is $O(n \cdot d)$. Finally, the normalized time mixing result $\mathbf{F}_B$ is shown in Exp (9):

$$\begin{aligned}\mathbf{F}_B &= \mathrm{LayerNorm}(\mathbf{T}) \\ &= \gamma \cdot \frac{\mathbf{T} - \mu}{\sqrt{\sigma^2 + \epsilon}} + \beta \\ &= \gamma \cdot \frac{\mathbf{T} - \frac{1}{d}\sum_{j=1}^{d} \mathbf{T}_{i,j}}{\sqrt{\frac{1}{d}\sum_{j=1}^{d}\left(\mathbf{T}_{i,j}\right)^2 + \epsilon}} + \beta,\end{aligned} \tag{9}$$

where $\gamma$ and $\beta$ are learnable parameters. The computational complexity of Layer Norm mainly comes from the calculation of mean and variance and normalization processing. For each time step $i$, the computational complexity of the mean and variance is $O(n \cdot d)$, and the complexity of the normalization process is also $O(n \cdot d)$.

Overall, the total computational complexity of these three processes is $O(n \cdot d)$. Finally, the enhanced RWKV module uses a binary classifier with a binary cross-entropy function for backpropagation learning, as shown in Exp. (10):

$$\boldsymbol{\mathcal{L}}_{\mathrm{CE}} = -\frac{1}{N}\sum_{i=1}^{N}[y_i \log(\hat{y}_i) + (1 - y_i)\log(1 - \hat{y}_i)], \tag{10}$$

where $\boldsymbol{\mathcal{L}}_{\mathrm{CE}}$ is the loss function, $y_i$ is the true label, $\hat{y}_i$ is the predicted value, and $N$ is the number of samples.

(2) *QACM* module

After completing the training of the *AFED* module, to make the model lightweight and able to run on edge computing devices, the results of the *AFED* module training are transferred to the *QACM* module through knowledge distillation techniques.

The advantage of the QACM module lies in its use of knowledge distillation techniques to transfer the training results from the AFED module, making the model lightweight. This process significantly reduces the model's complexity, allowing it to run on edge computing devices and meet real-time processing requirements in resource-constrained environments. Furthermore, the QACM module effectively reduces computational costs while maintaining detection accuracy, achieving a good balance between performance and efficiency in the system.

Knowledge distillation is a model compression technique that transfers knowledge from the teacher model to the student model. To ensure the effectiveness of knowledge distillation, Bayesian optimization is used to select the most suitable temperature parameter *T*. In the TCVADS system, the teacher model is the *AFED* module, and the student model is the *QACM* module. The loss function for knowledge distillation consists of two parts: Hard Loss and Soft Loss. The Hard Loss $\boldsymbol{\mathcal{L}}_{\text{Hard}}$ is used to train the teacher model as shown in Exp (18) and the Soft Loss $\boldsymbol{\mathcal{L}}_{\text{Soft}}$ is used for knowledge distillation, transferring the soft labels from the teacher model to the student model, as shown in Exp (11):

$$\boldsymbol{\mathcal{L}}_{\text{Soft}} = \sum_{i=1}^{N} \boldsymbol{\mathcal{L}}_{\text{KL}}\left(\frac{z_i}{T}, \frac{\hat{z}_i}{T}\right) = \sum_{i=1}^{N}\left(\sum_j p_j \log\left(\frac{p_j}{q_j}\right)\right)\left(\frac{z_i}{T}, \frac{\hat{z}_i}{T}\right) \quad (11)$$

where $\boldsymbol{\mathcal{L}}_{\text{KL}}$ is the KL divergence loss function, $\boldsymbol{z}_i$ and $\hat{\boldsymbol{z}}_i$ are the predicted logits of the teacher model and student model, respectively, and $T$ is the temperature parameter. The outputs of the teacher model and student model are also influenced by $T$. For selecting the temperature $T$ in $\boldsymbol{\mathcal{L}}_{\text{Soft}}$, a Gaussian process regression model and Bayesian optimization are used to determine the optimal temperature parameter. Given the training dataset $\mathcal{D} = \{(T_i,\ \boldsymbol{\mathcal{L}}_{\text{Soft}}(T_i))\}_{i=1}^{n}$, the $\boldsymbol{\mathcal{L}}_{\text{Soft}}(T)$ is expected to follow the distribution described in $\boldsymbol{\mathcal{L}}_{\text{So}}\ (T) \sim \mathcal{GP}\big(\mu(T), k(T, T')\big)$, where $\mu(T)$ is the mean function, and $k(T, T')$ is the squared exponential kernel, as shown in $k(T, T') = \sigma_f^2 \exp\left(-\frac{(T-T')^2}{2l^2}\right)$, where $\sigma_f^2$ and $l$ are hyperparameters representing the signal variance and length scale, respectively. A set of initial temperature parameters $T_0$ is chosen, and the corresponding $\boldsymbol{\mathcal{L}}_{\text{Soft}}(T_0)$ is calculated. The initial samples are used to train the Gaussian process regression model. Based on the training dataset $\mathcal{D}$, For a new sample $T^*$, the predictive distribution can be represented in $\boldsymbol{\mathcal{L}}_{\text{Sof}}\ (T^*) |\ \mathcal{D}, T^* \sim \mathcal{N}\big(\mu(T^*), \sigma^2(T^*)\big)$.

Herein, $\boldsymbol{\mathcal{L}}_{\text{Soft}}(T)$ represents the target function values of the training samples. Based on the current Gaussian process model, the next candidate temperature parameter $T_{next}$ is chosen using the Expected Improvement (EI) sampling strategy, as shown in Exp. (12):

$$\begin{gathered} T_{next} = \arg\max \text{EI}(T) \\ \text{EI}(T) = \mathbb{E}\big[\max\big(0, \boldsymbol{\mathcal{L}}_{\text{Soft}}(T_{\text{best}}) - \boldsymbol{\mathcal{L}}_{\text{Soft}}(T)\big)\big], \end{gathered} \quad (12)$$

where $\boldsymbol{\mathcal{L}}_{\text{Soft}}(T_{\text{best}})$ is the soft loss of the current best temperature parameter. The Gaussian process model is continuously updated with new samples $(T_{next}, \boldsymbol{\mathcal{L}}_{\text{Soft}}(T))$ until the optimal temperature parameter $T_{opt}$ is found, as shown in Exp. (13):

$$T_{opt} = \text{argmin}_T \boldsymbol{\mathcal{L}}_{\text{Soft}}(T) \quad (13)$$

Ultimately, this ensures that the total loss function $\boldsymbol{\mathcal{L}}_{\text{Total}}$ converges to its minimum. The total loss $\boldsymbol{\mathcal{L}}_{\text{Total}}$ is expressed as shown in Exp. (14):

$$\boldsymbol{\mathcal{L}}_{\text{Total}} = \boldsymbol{\mathcal{L}}_{\text{Hard}} + \boldsymbol{\lambda}\boldsymbol{\mathcal{L}}_{\text{Soft}}(T_{opt}). \quad (14)$$

In this way, using Gaussian process regression models and Bayesian optimization techniques, the optimal temperature parameters can be effectively selected to minimize the total loss function and achieve efficient knowledge distillation from *AFED* to *QACM*. After the distillation process is completed, during the usage phase, only the fast interpretable feature extraction based on MobileNet is required, followed by real-time coarse-grained classification through the *QACM* module.

### *B. TCVADS fine-grained detection*

This section will introduce the fine-grained anomaly detection module of the proposed TCVADS system. After the coarse-grained detection, if the identified result indicates violence, the system will proceed to the second stage to identify the specific violent content in the video. Therefore, the primary task at this stage is to perform multi-class classification, identifying the specific type of violent behavior the anomalous input video segment belongs to, such as fighting, shooting, or verbal abuse, from a total of six anomaly categories.

To achieve more precise fine-grained detection, the design of the TCVADS system incorporates a cross-modal approach. In other words, the method utilizes both visual and textual modalities to identify fine-grained anomaly tasks. It is important to note that in real life, some street surveillance cameras do not have audio recording capabilities, and the recorded footage does not include text descriptions like YouTube videos. To address this issue, this paper proposes combining a pre-trained CLIP model with learnable prompts to generate a set of textual information, which is then aligned with the corresponding video for contrastive learning. This approach results in more accurate fine-grained detection. The generation of textual information follows these steps: first, CLIP is used to convert each frame of the video into text. Once basic sentence descriptions are obtained, the second step involves adding labels from the surveillance video, and the third step incorporates learnable prompts to construct textual information that aligns with the images.

The strength of the cross-modal approach lies in its ability to achieve more fine-grained anomaly detection by integrating both visual and textual modalities. This method not only overcomes the limitation of real surveillance scenes, which often lack audio and textual descriptions, but also enhances detection accuracy by using a pre-trained CLIP model and learnable prompts to generate textual information for contrastive learning with video data. Specifically, video frames are converted into text descriptions, and with the addition of surveillance labels and learnable prompts, the generated text is precisely aligned with the images. This improves the system's fine-grained detection capability, enabling TCVADS to more accurately identify subtle anomalies in complex real-world scenarios, thus enhancing overall detection performance.

Before explaining how to align video and text, it is important to emphasize that CLIP mainly consists of two encoders: an image encoder and a text encoder. Its primary function is to convert input images and text into vectors and maintain consistency between the vector representations of images and text through contrastive learning. This enables the vectors of images and text to query each other, achieving a thorough fusion of vector representations in both the image and text spaces.

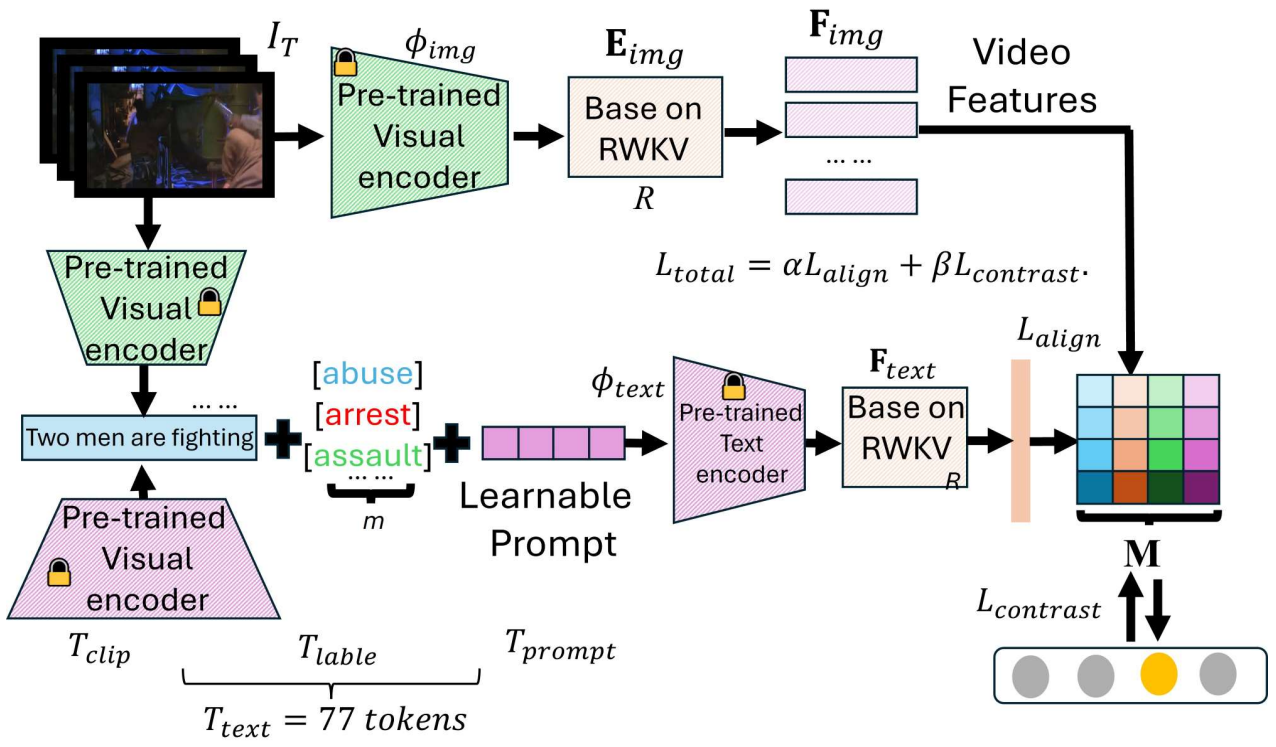

Fig. 3. The TCVADS Fine-Grained Working Flow.

Fig. 3 shows the Working Flow of fine-grained detection. It presents a framework which consists of image and text branches. As shown in Fig. 3, in the image branch, the CLIP is first used to extract the set of $T$ image features $\phi^{img} = \{\phi_1^{img}, \phi_2^{img}, \ldots, \phi_n^{img}\}$. In addition, the output of CLIP is a set of native texts $T^{CLIP}$=$\{T_1^{CLIP}, T_2^{CLIP}, \ldots, T_n^{CLIP}\}$. These features will be treated as the inputs of the enhanced RWKV module to handle long-sequence data, aiming to generate the video features $\boldsymbol{F}^{video} = \{\boldsymbol{F}_1^{video}, \boldsymbol{F}_2^{video}, \ldots, \boldsymbol{F}_m^{video}\}$ . In the text branch, the input of text branch consists of three parts: CLIP's native text, labels, and learnable prompts.

In the text branch, this paper introduces a novel input method to enhance the performance of the contrastive learning model. Firstly, for each frame $I_T$ of the video, a basic textual description $T^{clip}$ is obtained through the pre-trained CLIP model, which carries rich and universal visual semantic knowledge. Secondly, the corresponding original category labels $T^{label}$ (such as "fighting" or "robbery") from the dataset are directly encoded as tokens, ensuring that the model focuses on task-relevant semantic spaces. Lastly, a set of learnable tokens $T^{prompt}$ is introduced. These three types of information are combined as $\phi^{text}$ the input to CLIP's text encoder. Through the continuous optimization of these tokens during the training process, the model can specifically capture the subtle differences in anomalous events. These tokens provide actual classification information for the videos, helping the model understand the context before performing classification. CLIP can more accurately process and generate textual features, ensuring efficient and precise feature alignment in subsequent steps. In this research, the total length of $\phi^{text}$ is set to 77, which is the maximum length of CLIP's text encoder.

The text input with prompts is fed into the pre-trained text encoder, aiming to generate the set of $m$ text features $\phi^{text}$=$\{\phi_1^{text}, \phi_2^{text}, \ldots, \phi_n^{text}\}$.These features are then treated as the inputs of the enhanced RWKV model for generating features $\boldsymbol{F}^{tex} = \{\boldsymbol{F}_1^{text}, \boldsymbol{F}_2^{text}, \ldots, \boldsymbol{F}_m^{text}\}$. The alignment matrix $M_{m\times m} = [s_{i,j}]$ consists of the similarities between $m$ video features and $m$ text features, as shown in the following in

$$M_{m\times m} = \begin{matrix} & \boldsymbol{F}_1^{video} & \cdots & \boldsymbol{F}_m^{video} \\ \begin{matrix}\boldsymbol{F}_1^{text}\\ \vdots \\ \boldsymbol{F}_m^{text}\end{matrix} & \multicolumn{3}{c}{\begin{bmatrix} s_{1,1} & \cdots & s_{1,m} \\ \vdots & \ddots & \vdots \\ s_{m,1} & \cdots & s_{m,m} \end{bmatrix}} \end{matrix}.$$

The similarity value of $s_{i,j}$ in the i-th row and j-th column of matrix $M_{m\times m}$ can be calculated as shown in Exp. (15):

$$s_{i,j} = \frac{\boldsymbol{F}_i^{video} \cdot \boldsymbol{F}_j^{text}}{||\boldsymbol{F}_i^{video}|| \cdot ||\boldsymbol{F}_j^{text}||}, \tag{15}$$

Then, the similarity vector $S$, the prediction value $p_i$ or each category is calculated by Exp. (16) ：

$$p_i = \frac{\exp\,(s_{i,j}/\tau)}{\sum_j \exp\,(s_{i,j}/\tau)}, \tag{16}$$

where $\tau$ is a scaling hyper parameter. The loss function of the entire model includes the alignment loss $L_{align}$ and contrastive loss $L_{contrast}$. The alignment loss $L_{align}$ is shown in Exp. (17):

$$L_{align} = -\sum_{i=1}^{m} y_i \log(p_i) + \lambda_1 \left\|\theta_{align}\right\|^2, \tag{17}$$

where $y_i$ represents the labels, $\lambda_1$ controls the regularization of the model parameters in the alignment loss, and $\theta_{align}$ represents the model parameters related to the alignment loss.

The contrastive loss $L_{contrast}$ calculates the cosine similarity between normal class embeddings and other abnormal class embeddings, as shown in Exp. (18):

$$L_{contrast} = \sum_j \max\left(0, \frac{t_n . t_{aj}}{\|t_n\|_2 \cdot \left\|t_{aj}\right\|_2} - \delta\right) + \lambda_2 \|\theta_{contrast}\|^2, \tag{18}$$

where $t_n$ represents the embeddings of the normal class, $t_{aj}$ represents the embeddings of the abnormal class, $\delta$ is a hyperparameter , $\lambda_2$ is a regularization parameter, and $\theta_{contrast}$ represents the model parameters related to the contrastive loss. The final total loss function is shown in Exp. (19):

$$L_{total} = \alpha L_{align} + \beta L_{contrast}. \tag{19}$$

where $\alpha$ and $\beta$ are hyperparameters used to balance the weights of each loss term. Through this design, the image and text branches are aligned in the feature space, ensuring that each loss function plays a unique role during optimization, helping to avoid overfitting. During model training, the CLIP encoders are frozen, and the model is fine-tuned through backpropagation.

## V. Model Performance

In this section, relevant experimental performance and analysis are presented.

### A. Dataset

The experiments were conducted on two popular WSMAD datasets: UCF-Crime and XD-Violence. The dataset XD-Violence is currently the largest publicly available dataset in the violence detection domain. It comprises 4,754 videos, totaling 217 hours, and includes six types of violent events: verbal abuse, car accidents, explosions, fights, riots, and shootings. This dataset is randomly divided into a training set with 3,954 videos and a test set with 800 videos. The test set is further categorized into 500 violent videos and 300 non-violent videos. The UCF-Crime dataset consists of 1,900 real-world surveillance videos, with 1,610 for training and 290 for testing. It is important to note that the training videos in both the XD-Violence and UCF-Crime datasets only possess video-level labels.

### B. Implementation Details

The experimental parameters in this paper were set based on previous research [3][4][21]. The specific details are presented as follows:

In the fine-grained training, the proposed TCVADS adopts the pre-trained CLIP (ViT-B/16) but freezes its image and text encoders. The enhanced RWKV model is used for handling long-sequence data. The hyperparameters are set as follows: $\tau$ in Exp. (16) is 0.07, the context length $l$ is 40, and the sequence length in the RWKV module is 1024 for the XD-Violence dataset and 512 for the UCF-Crime dataset. The window length in the RWKV module is 256 for the XD-Violence dataset and 32 for the UCF-Crime dataset.

Regarding the hyperparameter settings, the $\tau$ in Exp. (16) is 0.08. The $\delta$ in Exp. (18) is 0.5. The The $\lambda_1$ in $L_{align}$ is $5 \times 10^{-4}$ for the XD-Violence dataset and $1 \times 10^{-3}$ for the UCF-Crime dataset. The $\lambda_2$ in $L_{contrast}$ is $6 \times 10^{-4}$ for the XD-Violence dataset and $2 \times 10^{-3}$ for the UCF-Crime dataset. For the hyperparameters and $\alpha$ and $\beta$ in Exp. (19), $\alpha$ is set to 1.2 and $\beta$ is set to 0.8. The hyperparameters $\gamma$ and $\beta$ in Exp. (9) are initialized to 1 and 0, respectively.

For the coarse-grained module, the improved MobileNet module is trained with the following training parameters: a learning rate of $3 \times 10^{-4}$, batch size of 256, and a total of 300 epochs.

TCVADS is trained on a single NVIDIA A100 GPU using PyTorch, with AdamW as the optimizer and a batch size of 256. For the XD-Violence dataset, the learning rate and total epochs are set to $1 \times 10^{-4}$ and 100, respectively. For the XD-Violence dataset, they are set to $6 \times 10^{-5}$ and 80, respectively.

### C. Comparison with State-of-the-Art Methods

TCVADS can simultaneously realize coarse-grained and fine-grained violence detection. The experiment presents the performance of the proposed TCVADS and compares it with several state-of-the-art methods on coarse-grained and fine-grained WSMAD tasks.

(1) *Coarse-grained WSMAD Results.*

Tables III and IV compare the performance of TCVADS, the method proposed in this paper, with other approaches for coarse-grained detection on the XD-Violence and UCF-Crime datasets. The results demonstrate that TCVADS performed well on both datasets, significantly outperforming other semi-supervised methods and classification-based weakly supervised approaches. Overall, TCVADS achieved state-of-the-art results with 85.58% AP on XD-Violence and 88.58% AUC on UCF-Crime. Compared to its strongest competitors, VadCLIP and CLIP-TSA, TCVADS improved the AP metric by 1.07% and 3.41% on XD-Violence and surpassed their AUC scores by 0.56% and 1.0% on UCF-Crime, respectively.

TABLE III
Coarse-grained comparisons on XD-Violence

| Category | Method | AP (%) |
|---|---|---|
| Semi supervised | SVM baseline | 50.80 |
| | OCSVM [38] | 28.63 |
| | C3D [6] | 31.25 |
| Weak supervised | CLIP [8] | 76.57 |
| | I3D [5] | 75.18 |
| | HD-Net [14] | 80.00 |
| | RTFM [12] | 78.27 |
| | AVVD [15] | 78.10 |
| | DMU [13] | 82.41 |
| | CLIP-TSA [4] | 82.17 |
| | VadCLIP [21] | 84.51 |
| | AnomalyCLIP [36] | 82.65 |
| | STPrompt [37] | 83.97 |
| | **TCVADS (Ours)** | **85.58** |

TABLE IV
Coarse-grained comparisons on UCF-Crime

| Category | Method | AUC (%) | Ano-AUC(%) |
|---|---|---|---|
| Semi-supervised | SVM baseline | 50.10 | 50.00 |
| | OCSVM [38] | 63.20 | 51.06 |
| | C3D [6] | 51.20 | 39.43 |
| Weak supervised | CLIP [8] | 84.72 | 62.60 |
| | I3D [5] | 84.14 | 63.29 |
| | HD-Net [14] | 84.57 | 62.21 |
| | AVVD [15] | 82.45 | 60.27 |
| | RTFM [12] | 85.66 | 63.86 |
| | DMU [13] | 86.75 | 68.62 |
| | UMIL [3] | 86.75 | 68.68 |
| | CLIP-TSA [4] | 87.58 | N/A |
| | VadCLIP [21] | 88.02 | 70.26 |
| | AnomalyCLIP [36] | 86.44 | 64.33 |
| | STPrompt [37] | 88.08 | 69.27 |
| | **TCVADS (Ours)** | **88.58** | **71.61** |

It's worth noting that AVVD, despite using multi-class labels for anomaly detection, only achieved 78.10% AP on XD-Violence and 82.45% AUC on UCF-Crime, significantly underperforming TCVADS. This suggests that simply increasing the number of labels doesn't necessarily improve anomaly detection performance; what matters more is how effectively these labels are utilized. Additionally, while VadCLIP [21], AnomalyCLIP [36], and STPrompt [37] introduced contrastive learning to enhance visual-language associations, their performance in coarse-grained tasks still fell short of TCVADS. This is because these methods use end-to-end architecture, which handles both coarse-grained and fine-grained detection tasks simultaneously, ultimately increasing the model's learning difficulty. The conflicting objectives of these tasks made it challenging for the optimization process to converge effectively, limiting the model's performance in coarse-grained detection.

In contrast, TCVADS employs a hierarchical design for coarse-grained and fine-grained tasks. For coarse-grained tasks, it uses an improved interpretable MobileNet module for feature extraction, ensuring fast and accurate results. It then employs an RWKV module for time series analysis and a knowledge distillation-optimized QACM module for identification on the time series. This approach extracts detailed information while avoiding the complexity of directly learning fine-grained labels. The modular, hierarchical design not only simplifies the learning process but also fully leverages complementary information at different levels, resulting in significantly improved performance.

(2) *Fine-grained WSMAD Results.*

For the performance of fine-grained WSMAD tasks, Tables V and VI respectively use the XD-Violence and UCF-Crime datasets to compare the effectiveness of the TCVADS developed in this paper with related studies. For fairness, all feature extractors use CLIP (ViT-B/16). Compared to coarse-grained experiments, the fine-grained task in WSMAD is more challenging as it requires not only detecting the occurrence of abnormal events but also accurately identifying specific abnormal categories.

As shown in Tables V and VI, the proposed TCVADS significantly outperforms other methods on the XD-Violence and UCF-Crime datasets. This improvement is mainly attributed to the innovative triplet input strategy in the TCVADS text encoder. Unlike other methods, such as VadCLIP [21], AnomalyCLIP [36], and STPrompt [37], which directly use text embeddings generated by CLIP or LLM, the proposed TCVADS cleverly integrates three complementary sources of information. First, for each video frame, textual descriptions are obtained through the pre-trained CLIP model and converted into tokens, carrying rich general visual semantic knowledge. Second, the original category labels from the dataset (e.g., "fighting" and "robbery") are directly encoded into tokens, ensuring that the model focuses on task-relevant semantic spaces. Lastly, a set of learnable tokens is introduced and continuously optimized during training to specifically capture subtle differences in abnormal events.

By combining these three types of tokens into the input sequence of the CLIP text encoder, TCVADS fully leverages the encoder's capacity and achieves an optimal integration of general semantics, task relevance, and dynamic subtlety features. This design allows TCVADS to more accurately match video content with abnormal event categories, thereby significantly outperforming other related methods in fine-grained classification tasks on the XD-Violence and UCF-Crime datasets.

*D. Ablation Study*

Experiments are conducted on the XD-Violence dataset with extensive ablation.

Table VII demonstrates the performance evaluation of the improved RWKV module proposed in this paper, compared with other time series modules such as Transformer and Transformer+GCN, on the XD-Violence dataset for both coarse-grained and fine-grained tasks. As shown in Table VII, time series modules are crucial for improving model performance. The baseline model without time series modeling performed the worst, with an AP of only 73.25% and an AVG of only 16.68%. Using a global Transformer encoder significantly improved the AP to 83.04%, but the AVG improvement was limited (17.26%). In contrast, the local Transformer encoder achieved AP and AVG values of 81.85% and 18.94%, respectively. This is because the local Transformer encoder can more effectively capture short-term dependencies, which is crucial for fine-grained analysis, as it requires accurate identification of subtle changes in consecutive frames.

The method using GCN alone performed well in both AP

TABLE V

FINE-GRAINED COMPARISONS ON XD-VIOLENCE

| Method | mAP@IOU(%) | | | | | |
|---|---|---|---|---|---|---|
| | 0.1 | 0.2 | 0.3 | 0.4 | 0.5 | AVG |
| Random Baseline | 1.82 | 0.92 | 0.48 | 0.23 | 0.09 | 0.71 |
| SVM | 18.64 | 12.53 | 11.05 | 8.26 | 4.32 | 10.96 |
| OCSVM [38] | 19.85 | 9.06 | 8.56 | 6.55 | 6.03 | 10.01 |
| C3D [6] | 20.28 | 13.72 | 8.44 | 5.06 | 2.81 | 10.06 |
| I3D [8] | 22.72 | 15.57 | 9.98 | 6.20 | 6.78 | 12.25 |
| HD-Net [14] | 35.35 | 28.02 | 20.94 | 15.01 | 10.33 | 21.93 |
| AVVD [15] | 30.51 | 25.75 | 20.18 | 14.83 | 9.79 | 20.21 |
| RTFM [12] | 31.25 | 26.85 | 21.94 | 13.56 | 12.54 | 21.23 |
| DMU [13] | 32.33 | 28.88 | 22.57 | 14.33 | 13.68 | 22.36 |
| UMIL [3] | 34.44 | 27.13 | 22.63 | **19.85** | 13.24 | 23.46 |
| CLIP-TSA [4] | 34.53 | 32.88 | 28.11 | 13.65 | 10.01 | 23.84 |
| VadCLIP [21] | 37.03 | 30.84 | 23.38 | 17.90 | 14.31 | 24.70 |
| AnomalyCLIP [36] | 36.83 | 26.31 | 22.80 | 18.63 | 12.56 | 23.42 |
| STPrompt [37] | 38.21 | 25.63 | **28.66** | 13.11 | 11.63 | 23.44 |
| **TCVADS (Ours)** | **39.58** | **31.26** | 28.35 | 19.28 | **16.29** | **26.95** |

TABLE VI

FINE-GRAINED COMPARISONS ON UCF-CRIME

| Method | mAP@IOU(%) | | | | | |
|---|---|---|---|---|---|---|
| | 0.1 | 0.2 | 0.3 | 0.4 | 0.5 | AVG |
| Random Baseline | 0.21 | 0.14 | 0.04 | 0.02 | 0.01 | 0.08 |
| SVM | 3.64 | 3.59 | 2.39 | 1.34 | 0.98 | 2.39 |
| OCSVM [38] | 4.85 | 4.44 | 2.33 | 1.85 | 1.56 | 3.01 |
| C3D [6] | 4.33 | 3.88 | 2.46 | 1.66 | 2.11 | 2.89 |
| I3D [8] | 5.73 | 4.41 | 2.69 | 1.93 | 1.44 | 3.24 |
| HD-Net [14] | 8.47 | 6.88 | 5.62 | 4.91 | 3.38 | 5.85 |
| AVVD [15] | 10.27 | 7.01 | 6.25 | 3.42 | 3.29 | 6.05 |
| RTFM [12] | 12.59 | 7.54 | 6.44 | 5.42 | 1.54 | 6.71 |
| DMU [13] | 11.32 | 7.62 | 5.97 | 4.33 | 2.36 | 6.32 |
| UMIL [3] | 11.84 | 7.85 | 6.52 | 3.97 | 2.84 | 6.60 |
| CLIP-TSA [4] | **12.62** | **8.13** | 6.66 | 4.28 | 1.91 | 6.72 |
| VadCLIP [21] | 11.72 | 7.83 | 6.40 | 4.53 | 2.93 | 6.68 |
| AnomalyCLIP [36] | 9.88 | 6.49 | 6.28 | 4.65 | 1.74 | 5.80 |
| STPrompt [37] | 11.56 | 7.49 | 6.13 | 5.11 | 2.11 | 6.48 |
| **TCVADS (Ours)** | 12.58 | 8.12 | **6.66** | **5.29** | **3.53** | **7.24** |

and AVG, achieving performances of 82.57% and 23.81% respectively, highlighting the advantages of graph structures in capturing spatiotemporal relationships. However, the direct combination of global and local Transformers performed poorly, possibly due to redundancy or interference between the two attention mechanisms, leading to degraded model performance. Interestingly, when combined separately with GCN, both global Transformer and local Transformer achieved significant improvements. The combination of global Transformer and GCN performed excellently in both AP and AVG, reaching 85.53% and 21.84% respectively, while the combination of local Transformer and GCN achieved the highest performance in AVG with a value of 24.76%, and an AP performance of 85.13%. This is mainly because GCN can effectively capture spatiotemporal dependencies between video frames, while Transformer excels in capturing global or local attention mechanisms. By combining GCN with the Transformer, the graph structure advantages of GCN and the attention mechanism of the Transformer can be fully utilized, thus achieving better spatiotemporal feature fusion and anomaly detection effects.

Traditional RWKV modules use two components, including TimeMixer and Channel Mixer. TimeMixer is mainly used for time series analysis, while Channel Mixer represents spatial fusion between different channels. Experiments have shown that in the WSMAD task, the Channel Mixer module does not improve model performance. This is because the Channel Mixer module is mainly responsible for fusing information between different channels, which in some cases may introduce unnecessary noise, interfering with the model's capture of spatial information. Furthermore, the TCVADS proposed in this paper has already completed spatial feature modeling in the first-stage feature extraction module, therefore the channel fusion operations in the time series module may lead to information redundancy, adversely affecting the overall performance of the model.

TABLE VII

EFFECTIVENESS OF RWKV

| Method | AP (%) | AVG (%) |
|---|---|---|
| Baseline (no temporal modeling) | 73.25 | 16.68 |
| Global Transformer | 83.04 | 17.26 |
| Local Transformer | 81.85 | 18.94 |
| Only GCN | 82.57 | 23.81 |
| Local Transformer + Global Transformer | 79.96 | 19.98 |
| Global Transformer + GCN | 85.53 | 21.84 |
| Local Transformer + GCN (LGT-Adapater) | 85.13 | 24.76 |
| RWKV | 84.27 | 24.71 |
| **The enhanced RWKV** | **85.58** | **26.95** |

The enhanced RWKV module proposed in this paper differs in that it uses two TimeMixer blocks, ensuring that the RWKV module focuses on time series analysis. Experimental results show that the enhanced RWKV proposed in this study performs most prominently, with an AP of 85.58% and an AVG as high as 26.95% in terms of performance. Compared to the traditional RWKV module, which has an AP of 84.27% and an AVG of 24.71%, the improved RWKV module proposed in this paper outperforms the traditional RWKV. The main reason is that the improved RWKV framework proposed in this study maintains high efficiency and stability through two TimeMixer blocks, while also achieving high accuracy in detecting short-term anomalous events.

Table VIII demonstrates a performance comparison of different prompting methods in the CLIP text encoder for fine-grained tasks. This experiment compared the TCVADS method proposed in this paper, handcrafted prompts, learnable prompts [35], frame-level average prompts [14], and anomaly-specific prompts [21]. The experimental results show that the proposed TCVADS method performed most excellently. The main reason lies in the input design of the TCVADS method for the text encoder, which cleverly integrates learnable prompts, original category labels, and CLIP visual prompts, forming an information-rich and highly targeted input sequence.

The main spirit of this research's approach is to maximize the utilization of knowledge from pre-trained models and optimize for specific tasks by fusing multiple prompting methods. Compared to traditional handcrafted prompts, TCVADS improved the AVG metric by 8.49%. This significant improvement highlights TCVADS's ability to apply the pre-trained knowledge of large language-vision models to the WSMAD task, thereby enhancing the performance of this task. Meanwhile, in comparison with simply using visual prompts, TCVADS's AVG metric is 5.42% higher, demonstrating its more effective use of visual information. Notably, TCVADS not only surpasses methods using learnable prompts alone, improving the AVG performance by 1.64%, but it also outperforms the combination of labels and learnable prompts, increasing the AVG performance by 1.69%.

TABLE VIII

EFFECTIVENESS OF PROMPT

| Prompt | AVG(%) |
|---|---|
| Hand-crafted Prompt | 18.46 (-8.49) |
| Learnable-Prompt | 25.31 (-1.64) |
| Average-Frame Visual Prompt | 21.53 (-5.42) |
| Anomlay-Focus Visual Prompt | 25.26 (-1.69) |
| **TCVADS (Our Method)** | **26.95** |

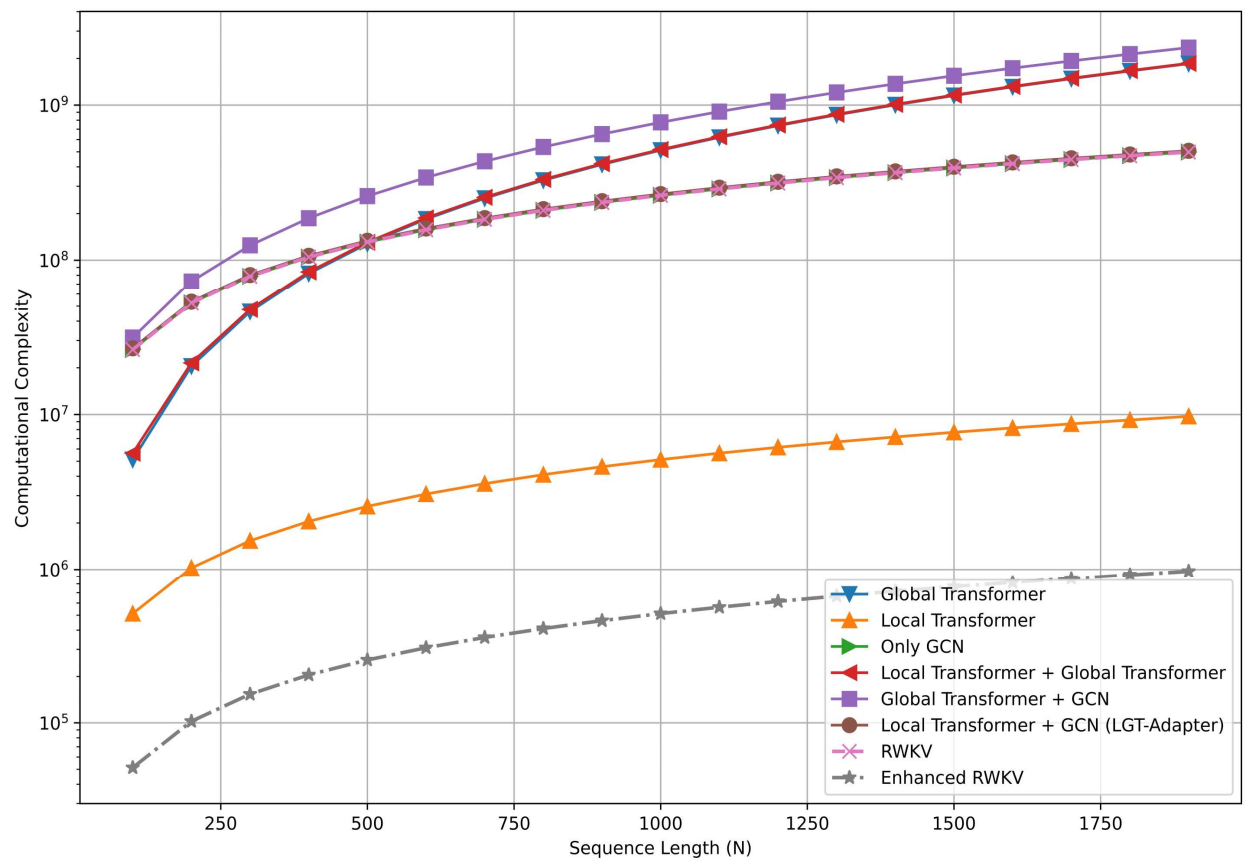

Fig.4. Comparison of Complexity Statistics for Different Methods

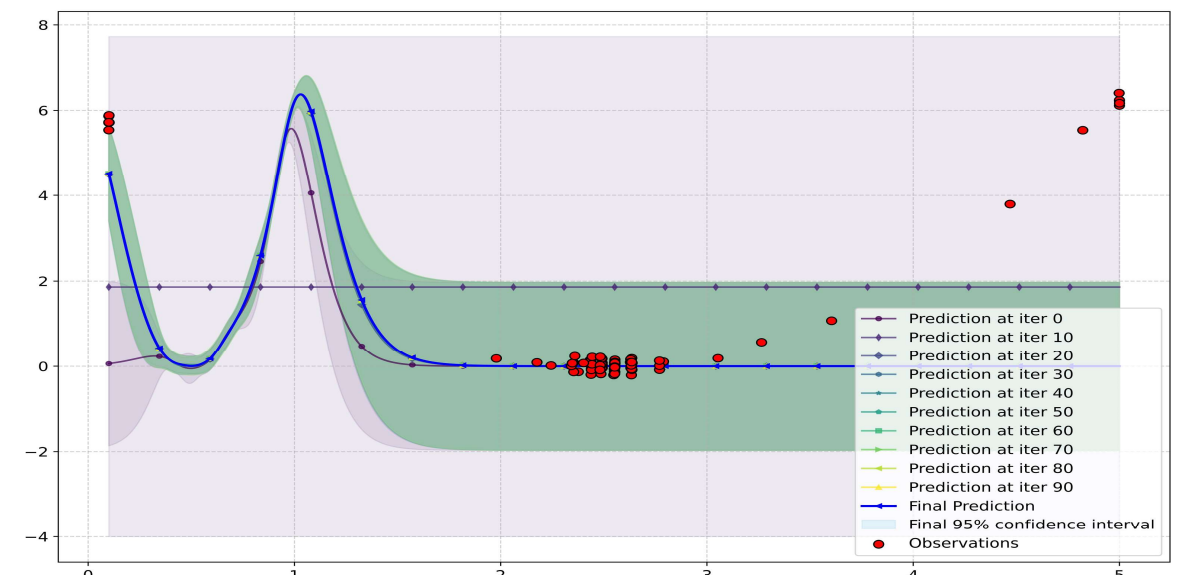

Fig. 5. Temperature Variation for Knowledge Distillation Using Bayesian Optimization

Recall that the TCVADS proposed in this paper includes CLIP's original input, labels, and learnable prompts in its input. These consistent and significant performance improvements demonstrate the effectiveness of TCVADS and its prompt combination strategy. It not only better adapts to and utilizes the pre-trained knowledge of large models but is also particularly adept at capturing features of anomalous video segments. This precise focus on anomalous instances enables TCVADS to generate more accurate instance-specific text representations, thereby achieving breakthrough progress in video-language alignment tasks.

Fig. 4 aims to compare the computational complexity of our proposed improved RWKV model with that of other related studies. In Fig. 4, the horizontal axis represents the sequence length (N), varying from 250 to 1750. The vertical axis represents complexity. This experiment helps to more clearly compare the performance differences between different models.

As shown in Fig. 4, the modules developed based on the Transformer model [15] exhibit higher initial complexity. However, their complexity increases exponentially with sequence length due to the model's stacked structure of the Transformer. This results in models such as Local Transformer + GCN (LGT-Adapter) [13] performing excellently in terms of model effectiveness, but with relatively low operational efficiency. In contrast, the improved RWKV module proposed in this paper demonstrates the lowest computational complexity. Although its complexity also increases linearly with the sequence length, the rate of increase is significantly lower. This characteristic makes the improved RWKV model particularly outstanding when processing long sequence tasks in video detection and deploying to edge devices.

Recall that this paper uses Bayesian optimization to select the temperature parameter $T$ for knowledge distillation. In Fig. 5, the horizontal axis represents the temperature parameter $T$, ranging from 0 to 5, while the vertical axis represents the value of the knowledge distillation loss function $\mathcal{L}_{\text{Total}}$, ranging from -4 to 6. The blue curves represent the changes in the objective function's loss value under different iterations of the Bayesian optimization algorithm, starting from 0 iterations, with a curve drawn every 10 iterations, up to 80. The thickest blue line is the final decision. The green shaded area represents the 95% confidence interval of the Bayesian optimization algorithm for the objective function value, and the red dots are the actual observed values at different $T$ values.

As shown in Fig. 5, during the Bayesian optimization process, the objective function of knowledge distillation changes with the number of iterations. As the algorithm samples more points and observes actual performance, the variation curve of the $\mathcal{L}_{\text{Total}}$ gradually forms a complex "W" shape, eventually converging to a stable double-peak structure. This process reflects how Bayesian optimization gradually constructs a probability model of the parameter space through sequential sampling.

The double-peak structure of the curve (at T≈1 and T≈2.5) reveals an interesting phenomenon: knowledge distillation performance may reach local optimal in two different temperature regions. The global optimum appears at T≈2.5, where the predicted objective function value is almost 0, indicating that Bayesian optimization successfully found the optimal temperature parameter for knowledge distillation.

The change in confidence intervals provides crucial information about the algorithm's uncertainty. In the regions where $T<1$ and $T>3$, the confidence intervals are wider, indicating higher prediction uncertainty in these areas, possibly due to fewer observation points or large variations in observed values. In contrast, for $1<T<3$, especially around $T≈2.5$, the confidence interval significantly converges, indicating that Bayesian optimization is very confident in its predictions in this region. Most of the actual observation points (red dots) are concentrated in this region and close to the prediction curve, further verifying the accuracy of Bayesian optimization.

The experimental results demonstrate the power of using Bayesian optimization in automatically selecting the temperature parameter for knowledge distillation. It not only precisely locates the optimal parameter (T≈2.5) but also reveals how the algorithm efficiently explores the parameter space and quantifies uncertainty through the evolution of confidence intervals and prediction curves. This method not only eliminates the tedious process of manual parameter tuning but also provides comprehensive insights into how parameters affect model performance.

### *E. Interpretable and Qualitative Analyses*

This section introduces the interpretability experiments for each module in the TCVADS system proposed in this study. For coarse-grained interpretability experiments, the CAM method [22] is used to verify the interpretability of cases. For fine-grained interpretability experiments, t-SNE [23] is used to reduce the dimensionality and visualize the high-dimensional vectors in CLIP.

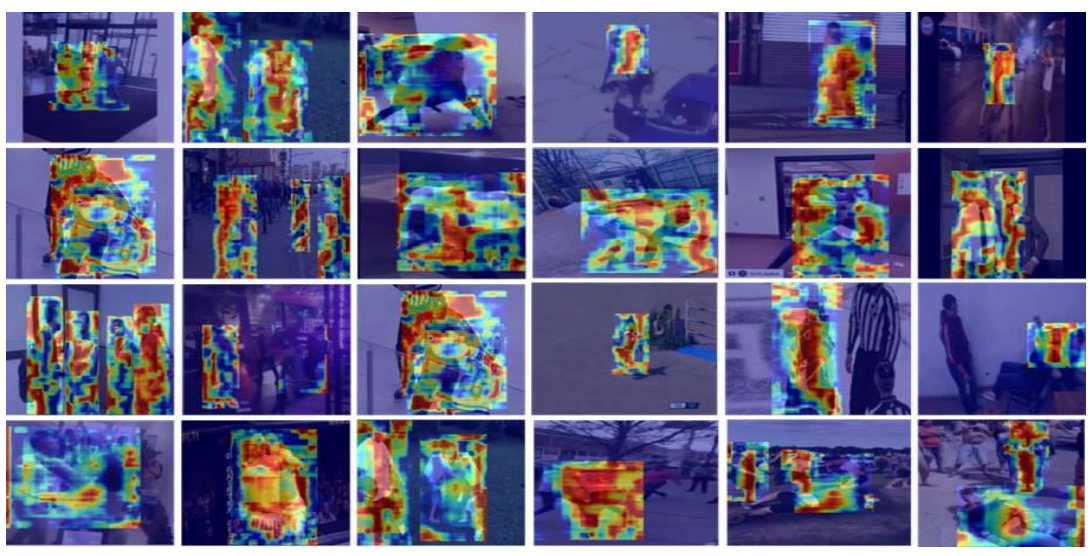

Fig. 6. Interpretable Analysis Based on the MobileNet Module.

Fig.6 visually demonstrates the decision-making process of the MobileNet module which is improved by TCVADS. The experiment focuses on the specially designed deconvolution layer in MobileNet for visualization purposes. To comprehensively evaluate the model's judgment capabilities, a series of representative fighting videos were selected from the test dataset. Firstly, the features extracted by the MobileNet module are converted into heat maps. As shown in Fig. 6, the attention of heatmaps is highly concentrated on actual violent behaviors, such as punching, kicking, or shoving actions. These hotspot areas highly correspond with the video sections that human observers would identify as abnormal or dangerous, indicating that TCVADS can not only accurately detect anomalous events but also precisely locate specific actions or interactions when abnormal behavior occurs.

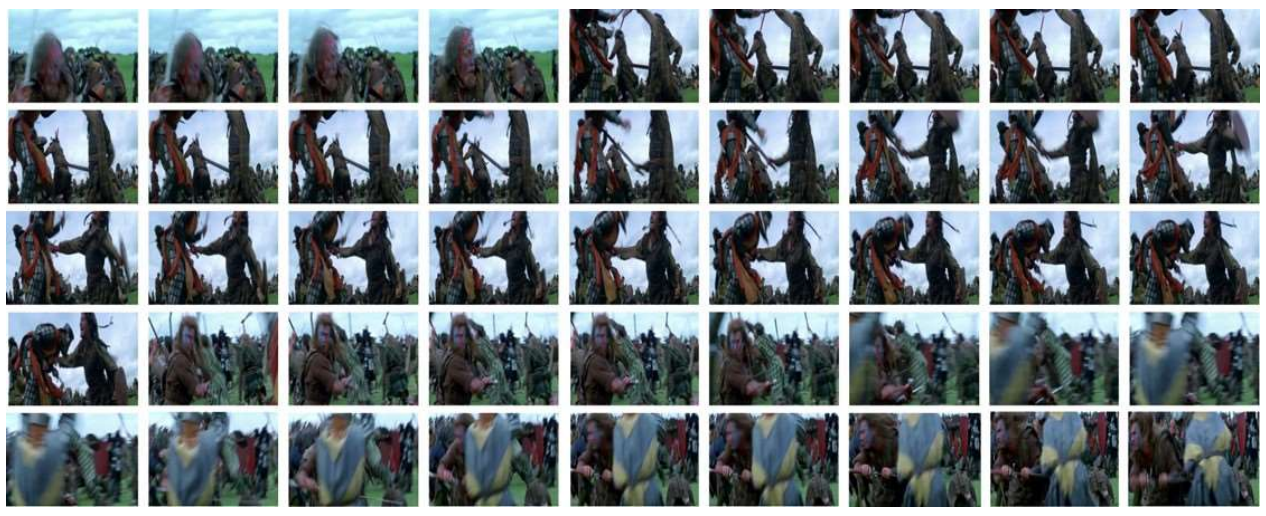

(a). Abnormal video clips.

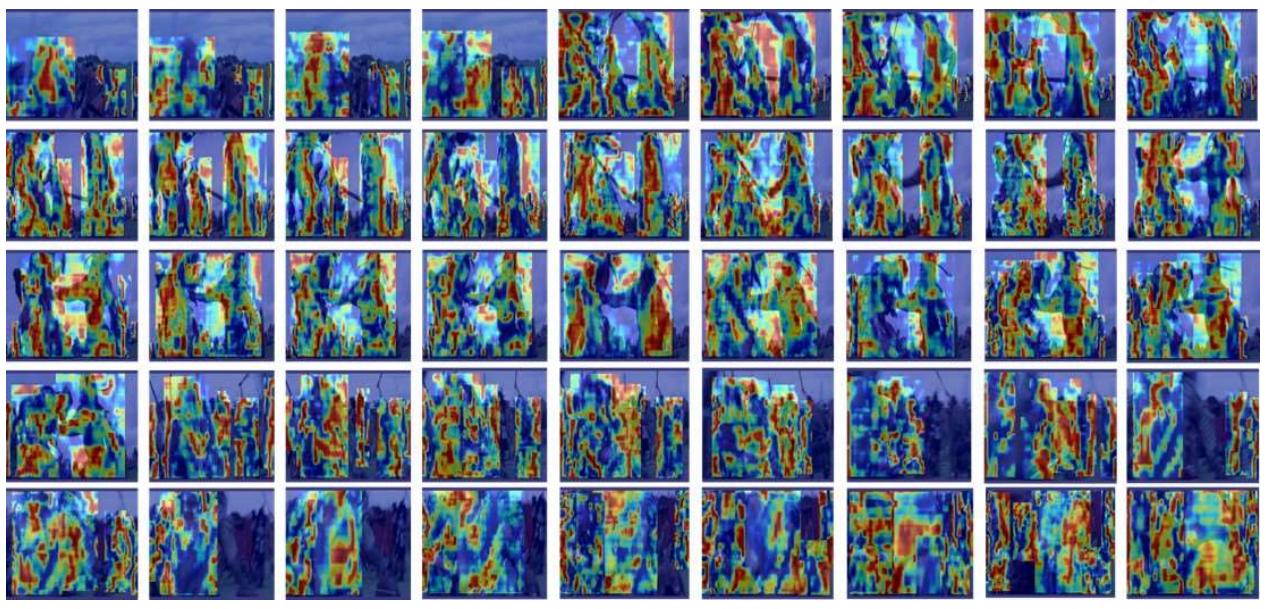

(b). Interpretable heat map before knowledge distillation.

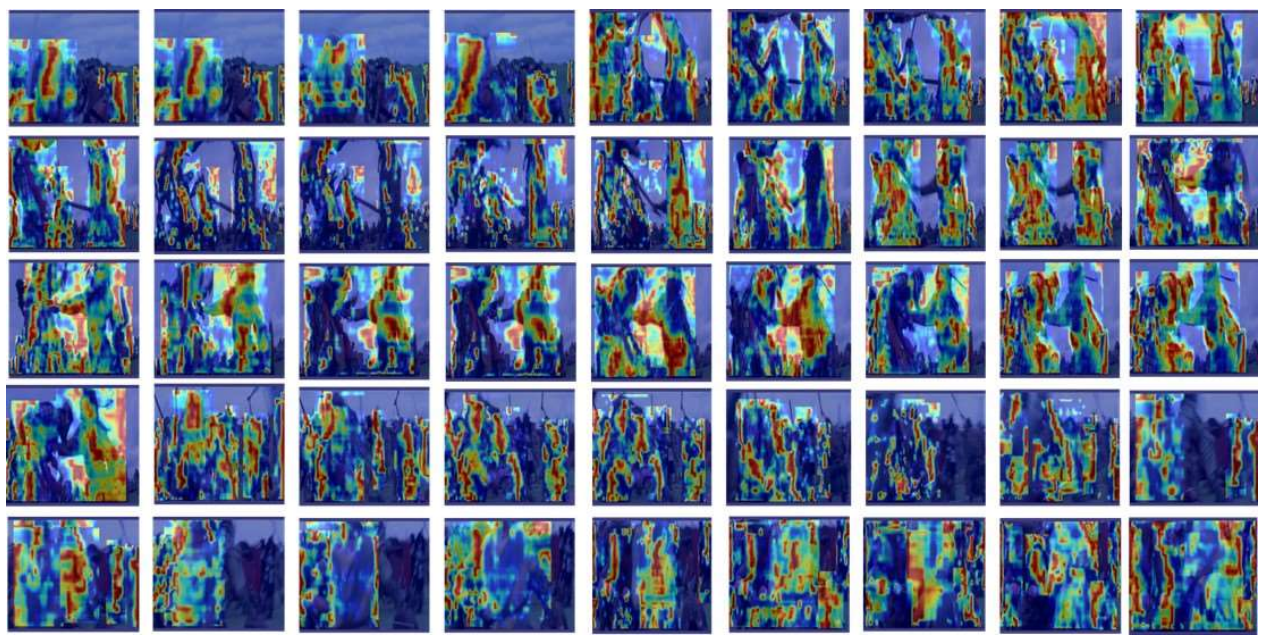

(c). Interpretable heat map after knowledge distillation.

Fig. 7. Interpretable Analysis knowledge distillation.

Fig.7 aims to illustrate the interpretability of the model before and after knowledge distillation. The Fig. is divided into three subFig. s. Fig. 7(a) shows a sequence of video frames containing abnormal behavior, possibly depicting a fighting scene. Let $HM^{before}$ and $HM^{after}$ denote the heatmaps before and after applying knowledge distillation technique, respectively. Figs. 7(b) and 7(c) show the heatmaps of $HM^{before}$ and $HM^{after}$, respectively.

As shown in Fig. 7(b), the $HM^{before}$ are more dispersed, with the model's attention less focused. In contrast, the $HM^{after}$, as shown in Fig. 7(c), show the heatmaps with a more concentrated and clearer distribution of the model's attention, enabling the proposed TCVADS to more accurately pinpoint abnormal or important areas. This indicates that knowledge distillation has enhanced the model's interpretability, making the decision-making process more transparent and reliable.

Fig. 8 aims to use t-SNE [23] to perform interpretable visual analysis of feature distributions in XD-Violence, where asterisks represent text label features. Each color corresponds to a different anomaly type in XD-Violence. Fig. 8 displays visualization analyses of related studies CLIP [8], AVVD [14], VadCLIP [21], and the TCVADS proposed in this research.

As shown in Fig. 8, CLIP already demonstrates general capabilities based on image-text pairs. However, it still cannot effectively distinguish between different categories in the WSMAD task. In the related study AVVD, the visual features can better align with text features. For VadCLIP, it also shows some advantages in enhancing feature recognition ability. By integrating multimodal information, it forms a tighter fusion between visual and textual features.

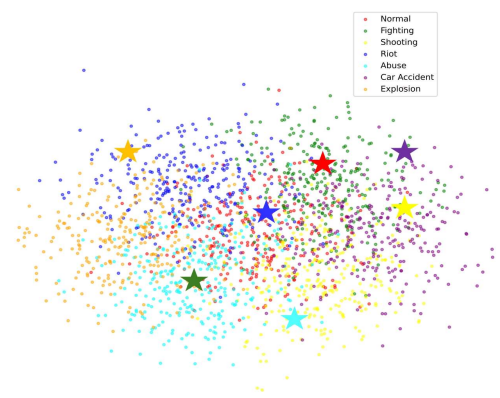

(a). CLIP

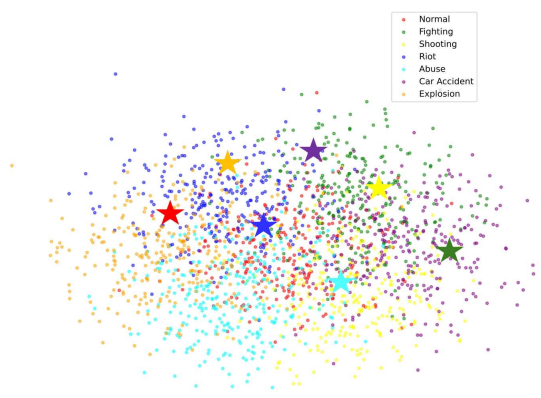

(b). AVVD

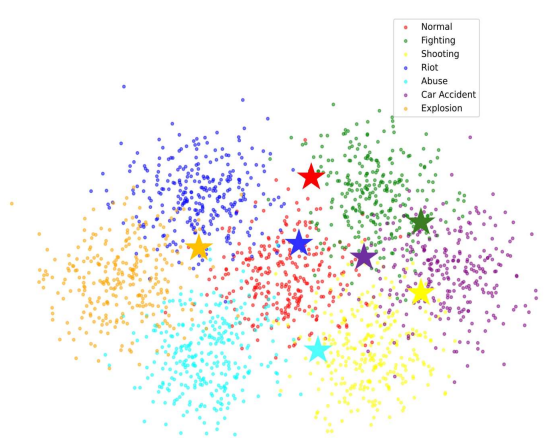

(c). VadCLIP

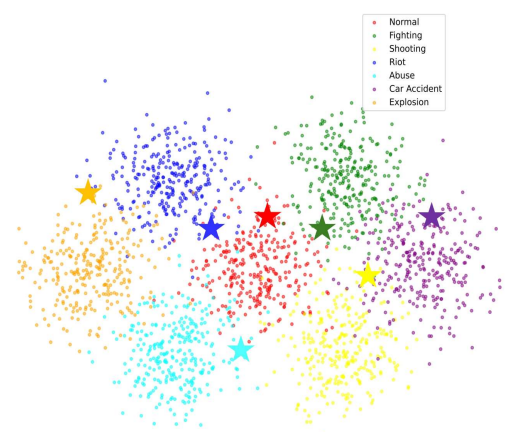

(d). TCVADS

Fig.8. t-SNE [23] visualizations.

Nevertheless, the proposed TCVADS model shows the most significant advantages. As shown in Fig. 8 (d), the TCVADS not only more effectively distinguishes different anomalous video categories but also shows a more concentrated distribution of feature points. This is because TCVADS, in its fine-grained detection design, adopts a unique ternary relationship combination for the text encoder input, combining the original CLIP text, labels, and learnable prompts. This enables better alignment of text and images, which also reduces the scattered distribution of feature points, allowing subsequent clustering or classification tasks to achieve better performance.

This indicates that TCVADS can more precisely capture and characterize features of different categories, not only improving classification accuracy but also enhancing the robustness and interpretability of models, demonstrating superior performance in handling complex video anomaly detection tasks.

TABLE IX

THE INTERPRETABLE QUANTITATIVE EVALUATION

| Related Work | $R^2$ ↑ |
|---|---|
| SVM baseline | 23.4 |
| OCSVM [38] | 27.1 |
| C3D [6] | 13.8 |
| CLIP [8] | 22.8 |
| I3D [5] | 19.3 |
| HD-Net [13] | 14.3 |
| RTFM [12] | 31.8 |
| AVVD [14] | 38.2 |
| DMU [13] | 27.4 |
| CLIP-TSA [4] | 36.6 |
| VadCLIP [21] | 45.2 |
| STPrompt [36] | 38.8 |
| AnomalyCLIP [37] | 51.8 |
| **TCVADS (Coarse-grained)** | **64.8** |
| **TCVADS (Fine-grained)** | **53.7** |

Table IX presents the quantitative evaluation metrics for the interpretability experiment. This paper builds methods from related studies [39] [40], utilizing the interpretable machine learning technique LIME [41] to refit the predicted values of the test samples and compare the refitted curve to the original model's predicted curve to assess the $R^2$ value. A higher $R^2$ value indicates greater interpretability of the model, while a lower $R^2$ value suggests weaker interpretability. The dataset used for the interpretability experiment is XD-Violence.

As shown in Table IX, the proposed TCVADS two-stage system achieves higher $R^2$ values in both the coarse-grained and fine-grained stages compared to other related methods. This suggests that the curves refitted through interpretable machine learning are closer to the original model's predictions, further demonstrating that the TCVADS system model has greater interpretability.

TABLE X

COMPREHENSIVE EVALUATION

| Related Work | Feature | Parameter | FLOPS |
|---|---|---|---|
| C3D [6] | C3D | 78M | 386.2G |
| RTFM [12] | I3D | 28M | 186.9G |
| AnomalyCLIP [36] | Clip | 45.68M | 85.8G |
| STPrompt [37] | Clip | 31.5M | 44.8G |
| **TCVADS (Coarse-grained)** | MobileNet | 3.6M | 0.576MB |
| **TCVADS (Fine-grained)** | Clip | 28.4M | 22.17G |

TABLE X aims to present the parameters (Parameter) and computational complexity (FLOPS) of the proposed TCVADS, C3D, RTFM, AnomalyCLIP, and STPrompt on the XD-Violence dataset. Parameter represents the number of weights in the model, used to measure the model's complexity and storage requirements, while FLOPS represents the computational load required for a single forward pass. As shown in Table X, the proposed TCVADS uses MobileNet-based feature extraction for coarse-grained feature extraction, followed by the proposed enhanced RWKV module to analyze the temporal sequence, concluding with binary classification. This lightweight design significantly reduces both the model's parameters and the computational cost of a single forward pass compared to other approaches. Moreover, TCVADS employs the enhanced RWKV module in its fine-grained design as well, which allows it to achieve the lowest parameter count and computational load compared to other CLIP-based methods such as AnomalyCLIP and STPrompt. It is noteworthy that existing edge computing devices, such as the NVIDIA Jetson Nano, impose limits of 100M for model parameters and 100G FLOPS for computational load, while the Google Coral Edge TPU restricts model parameters to 32MB. The proposed TCVADS meets the operational requirements of both devices in both coarse-grained and fine-grained modes, making it more feasible for edge computing environments compared to other methods.

## VI . CONCLUSION

This paper proposes an interpretable Two-stage Cross-modal Video Anomaly Detection System (TCVADS). In the first stage, the coarse-grained detection stage, the proposed TCVADS first enhances the MobileNet module and improves the RWKV module, achieving rapid feature extraction and time series analysis, effectively reducing computational complexity and improving response time. Furthermore, in the first stage, to further enhance real-time performance and interpretability, this system adopts knowledge distillation techniques, transferring the coarse-grained task to the QACM module, thus enabling rapid detection and demonstrating its interpretability capabilities in edge computing environments. In the second stage, called the fine-grained detection stage, TCVADS adopts an innovative ternary input strategy for the contrastive learning text encoder, better adapting and utilizing pre-trained knowledge from large models to improve detection accuracy and interpretability.

Using XD-Violence and UCF-Crime as experimental datasets, the results show that the proposed TCVADS significantly outperforms existing methods in terms of AP and AUC evaluation metrics. Moreover, the system demonstrates powerful interpretability and shows great application potential in areas such as intelligent surveillance and traffic monitoring. Future research will further optimize cross-modal contrastive learning methods and explore broader application scenarios to enhance the system's robustness and generalization capabilities.

## *Authors' Biographies*

Wen-Dong Jiang (*Graduate Student Member*, *IEEE*) received his Bachelor's degree in Multimedia and Game Science Development from Lunghwa University of Science and Technology, Taoyuan, Taiwan, in 2021. In 2022, he studied Software Engineering at Fayette Institute of Technology, Oak Hill, West Virginia, USA. He obtained his master's degree in computer science and information engineering from Ming Chuan University, Taoyuan, Taiwan, in 2023. Currently, he is pursuing a Ph.D. in Computer Science and Information Engineering at Tamkang University, Taiwan. His research interests include interpretable machine learning and AIoT.

Chih-Yung Chang (*Member, IEEE*) received the Ph.D. degree in computer science and information engineering from the National Central University, Zhongli, Taiwan, in 1995. He is currently a Full Professor with the Department of Computer Science and Information Engineering, Tamkang University, New Taipei City, Taiwan. His current research interests include internet of things, wireless sensor networks, ad hoc wireless networks, and Long Term Evolution (LTE) broadband technologies. He has served as an Associate Guest Editor for several SCI-indexed journals, including the *International Journal of Ad Hoc and Ubiquitous Computing* from 2011 to 2024, *the International Journal of Distributed Sensor Networks* from 2012 to 2024, *IET Communications* in 2011, *Telecommunication Systems* in 2010, the *Journal of Information Science and Engineering* in 2008, and the *Journal of Internet Technology* from 2004 to 2018.

Hsiang-Chuan Chang received his Ph.D. from the Department of Transportation Management at Waseda University, Japan, in 2023. He is currently an Assistant Professor in the Department of Transportation Management at Tamkang University. His research interests include intelligent transportation systems, smart cities, and AIoT.

Ji-Yuan Chen received bachelor's degree in computer science and information engineering from Ming Chuan University, Taoyuan, Taiwan, Taoyuan, Taiwan, in 2022. Currently, he is pursuing a Ph.D. in the Faculty of Engineering and Information Technology, The University of Melbourne, Parkville VIC 3052, Australia. His research interests include interpretable machine learning and AIoT.

Diptendu Sinha Roy (*Senior Member, IEEE*) received the Ph.D. degree in engineering from the Birla Institute of Technology, Mesra, India, in 2010. In 2016, he joined the Department of Computer Science and Engineering, National Institute of Technology (NIT) Meghalaya, Shillong, India, as an Associate Professor, where he has been working as the Chair of the Department of Computer Science and Engineering since January 2017. Prior to his stint at NIT Meghalaya, he worked with the Department of Computer Science and Engineering, National Institute of Science and Technology, Berhampur, India. His current research interests include software reliability, distributed and cloud computing, and the Internet of Things (IoT), specifically applications of artificial intelligence/machine learning for smart integrated systems. Dr. Roy is a member of the IEEE Computer Society.